\documentclass[sigconf]{acmart}
\usepackage{subfigure}
\usepackage{appendix}
\usepackage{algorithm}
\usepackage{algorithmic}

\AtBeginDocument{%
  \providecommand\BibTeX{{%
    \normalfont B\kern-0.5em{\scshape i\kern-0.25em b}\kern-0.8em\TeX}}}

\acmConference[WWW '20]{Proceedings of The Web Conference 2020}{April 20--24, 2020}{Taipei, Taiwan}
\acmBooktitle{Proceedings of The Web Conference 2020 (WWW '20), April 20--24, 2020, Taipei, Taiwan}
\acmPrice{}
\acmDOI{10.1145/3366423.3380115}
\acmISBN{978-1-4503-7023-3/20/04}
\setcopyright{iw3c2w3}
\copyrightyear{2020}
\acmYear{2020}

\acmConference[WWW '20]{Proceedings of The Web Conference 2020}{April 20--24, 2020}{Taipei, Taiwan}
\acmBooktitle{Proceedings of The Web Conference 2020 (WWW '20), April 20--24, 2020, Taipei, Taiwan} 
\acmPrice{}
\acmISBN{978-1-4503-7023-3/20/04}
\begin{document}

\title{Hierarchical Adaptive Contextual Bandits for Resource Constraint based Recommendation}


\author{Mengyue Yang}
\authornote{This work is done during the first
author’s internship in AI Labs, Didi Chuxing. Qingyang Li is the corresponding author.}
\affiliation{
  \institution{University of Chinese Academy of Sciences}
  \city{Beijing, China}
}
\email{yangmengyue17@mails.ucas.edu.cn}

\author{Qingyang Li, Zhiwei Qin, Jieping Ye}
\affiliation{
  \institution{AI Labs, Didi Chuxing}
  \city{Mountain View, CA$/$Beijing, China}
}
\email{{qingyangli, qinzhiwei, yejieping}@didiglobal.com}

\renewcommand{\shortauthors}{Mengyue Yang and Qingyang Li, et al.}

\begin{abstract}
Contextual multi-armed bandit (MAB) achieves cutting-edge performance on a variety of problems. When it comes to real-world scenarios such as recommendation system and online advertising, however, it is essential to consider the resource consumption of exploration. 
%when maximizing the reward of bandit algorithms. 
In practice, there is typically non-zero cost associated with executing a recommendation (arm) in the environment, and hence, the policy should be learned with a fixed exploration cost constraint. 
It is challenging to learn a global optimal policy directly, since it is a NP-hard problem and significantly complicates the exploration and exploitation trade-off of bandit algorithms. 
Existing approaches focus on solving the problems by adopting the greedy policy which estimates the expected rewards and costs and uses a greedy selection based on each arm's expected reward/cost ratio using historical observation until the exploration resource is exhausted. 
However, existing methods are hard to extend to infinite time horizon, since the learning process will be terminated when there is no more resource. In this paper, we propose a hierarchical adaptive contextual bandit method (HATCH) to conduct the policy learning of contextual bandits with a budget constraint. HATCH adopts an adaptive method to allocate the exploration resource based on the remaining resource/time and the estimation of reward distribution among different user contexts. In addition, we utilize full of contextual feature information to find the best personalized recommendation. Finally, in order to prove the theoretical guarantee, we present a regret bound analysis and prove that HATCH achieves a regret bound as low as $O(\sqrt{T})$. The experimental results demonstrate the effectiveness and efficiency of the proposed method on both synthetic data sets and the real-world applications. 
\end{abstract}

\begin{CCSXML}
<ccs2012>
 <concept>
  <concept_id>10010520.10010575.10010755</concept_id>
  <concept_desc>Computing methodologies~Reinforcement learning</concept_desc>
  <concept_significance>300</concept_significance>
 </concept>
 <concept>
  <concept_id>10010520.10010553.10010562</concept_id>
  <concept_desc>Information systems~Recommender systems</concept_desc>
  <concept_significance>500</concept_significance>
 </concept>
</ccs2012>
\end{CCSXML}

\ccsdesc[300]{Computing methodologies~Reinforcement learning}
\ccsdesc[500]{Information systems~Recommender systems}

\keywords{reinforcement learning, recommendation system, contextual bandits, budget constrain, exploration resource allocation}


\maketitle

\section{Introduction}
The multi-armed bandit (MAB) is a typical sequential decision problem, in which an agent receives a random reward by playing one of K arms at each round and try to maximize its cumulative reward. 
%The agent is able to learn the inherent trade-off between exploration, identifying and learning the policy from an action, and exploitation, gathering accumulated reward from actions. 
A lot of real world applications can be modeled as MAB problems, such as news recommendation \cite{li2010contextual}\cite{zhou2016latent}\cite{shang2019environment}, auction mechanism design \cite{mohri2014optimal}, online advertising \cite{chou2014pseudo}\cite{slivkins2013dynamic}\cite{pandey2007bandits}\cite{li2016parallel}. Some works make full use of the observed $d$ dimension features associated with the bandit learning, referred to as contextual multi-armed bandits.

We focus on the bandit problems on the user recommendation under resource constraints. It is common in real-world scenarios such as online advertising and recommendation system.
%the position and impressions of the items and the quantity of display in a day are limited. 
Most of methods not only focus on improving the number of orders and clicks but also balance the exploration-exploitation trade-off within a limit exploration resource so that CTR(click/impression) and purchase rate are considered. Since the impressions of users are almost fixed in a certain scope(budget), it can be formulated as the problem of increasing the number of clicks under a budget scope. 
Thus, it is necessary to conduct the policy learning under constrained resources which indicates that cumulative displays of all items (arms) can not exceed a fixed budget within a given time horizon. In our scenarios, each action is treated as one recommendation and the total number of impressions as the limited budget. To enhance CTR, we treat every recommendation equally and formulate it as unit-cost for each arm. While, there exist some scenarios that cost can not be treated equally, such as advertising bidding. Advertising positions and recommendations are decided by dynamical pricing, which is associated with the field of Game Theory\cite{mohri2014optimal}. 

In this study, we aim at learning the policy to maximize the expected feedback like Click Through Rate (CTR) under exploration constraints. It can be formed as a constrained bandit problem. In such settings, the algorithm recommends an item (arm) for the incoming context in each round, and observes a feedback reward. Meanwhile, the execution of the action will produce the cost, a unit cost, which means exploration of policy learning brings resource consumption. There are a lot of studies on Resource-Constrained Bandits problem. \cite{badanidiyuru2013bandits} formulates a constraint multi-armed bandits as the bandits with knapsacks, which combines stochastic integer programming with online learning. And more recently, \cite{badanidiyuru2014resourceful} proposed a near-optimal MAB regret under the resource constraint combined with contextual setting of fixed resource and time-horizon bandits. \cite{balakrishnan2018using} proposed that when constraint is not time and resource, it considers the influences of humans behaviours during the policy learning. However, most of existing works focus on MAB problems in a finite state space, without considering user contextual information. Recently, \cite{agrawal2016linear} considered the linear contextual bandit problem and adopted a knapsack method while it requires a prior distribution of contextual information and costs depend on contexts but not arms. \cite{wu2015algorithms} proposes an adaptive linear programming to solve while it only deals with the UCB setting.

%To solve the resource constraint problems in contextual multi-armed bandit, it is necessary to optimize both the efficiency of learning policy and exploration strategy. It is challenging to obtain a global optimal policy since it is a NP-hard problem and we should consider not only remaining time horizon as well as the unknown upcoming user contextual information. Thus, the problem is converted to learn a more reasonable policy to balance the exploration and exploitation trade-offs of bandit algorithm with user contextual information.

In this paper, we propose a hierarchical adaptive learning structure to dynamically allocate the resource among different user contexts as well as to conduct the policy learning by making full use of the user contextual features. In our method, we not only consider the scale of resource allocation at the global level but the remaining time horizon. The hierarchical learning structure composes of two levels: the higher level is resource allocation level where the proposed method dynamically allocates the resource according to estimation of the user context value and the lower level is personalized recommendation level where our method makes full use of contextual information to conduct the policy learning alternatively. In summary, this study makes following contributions:
\begin{itemize}
\item We propose a novel adaptive resource allocation to balance the efficiency of policy learning and exploration resource under the remaining time horizon. To the best of our knowledge, this is the
first study to consider the dynamic resource allocation in the contextual multi-armed bandit problems.
\item In order to utilize the contextual information for users 
, we propose a hierarchical adaptive contextual bandits method (HATCH), estimating the reward distribution of user contexts to allocate the resources dynamically and employ user contextual features for personalized recommendation.
\item To demonstrate the regret guarantee of the proposed method, we present an analysis of regret bound and give a theoretically proof that HATCH has a regret bound of $O(\sqrt{T})$.
\end{itemize}
We define a bandit problem with limited resource and review some existing methods in Section 2. Then we propose a new framework HATCH and give a detailed description in Section 3. We give a regret analyze of HATCH in Section 4. Experiment results are shown in Section 5. Finally, we make a conclusion of this paper in Section 6.

\begin{figure}[!t]
{\includegraphics[height=1.3in,width=3.22in]{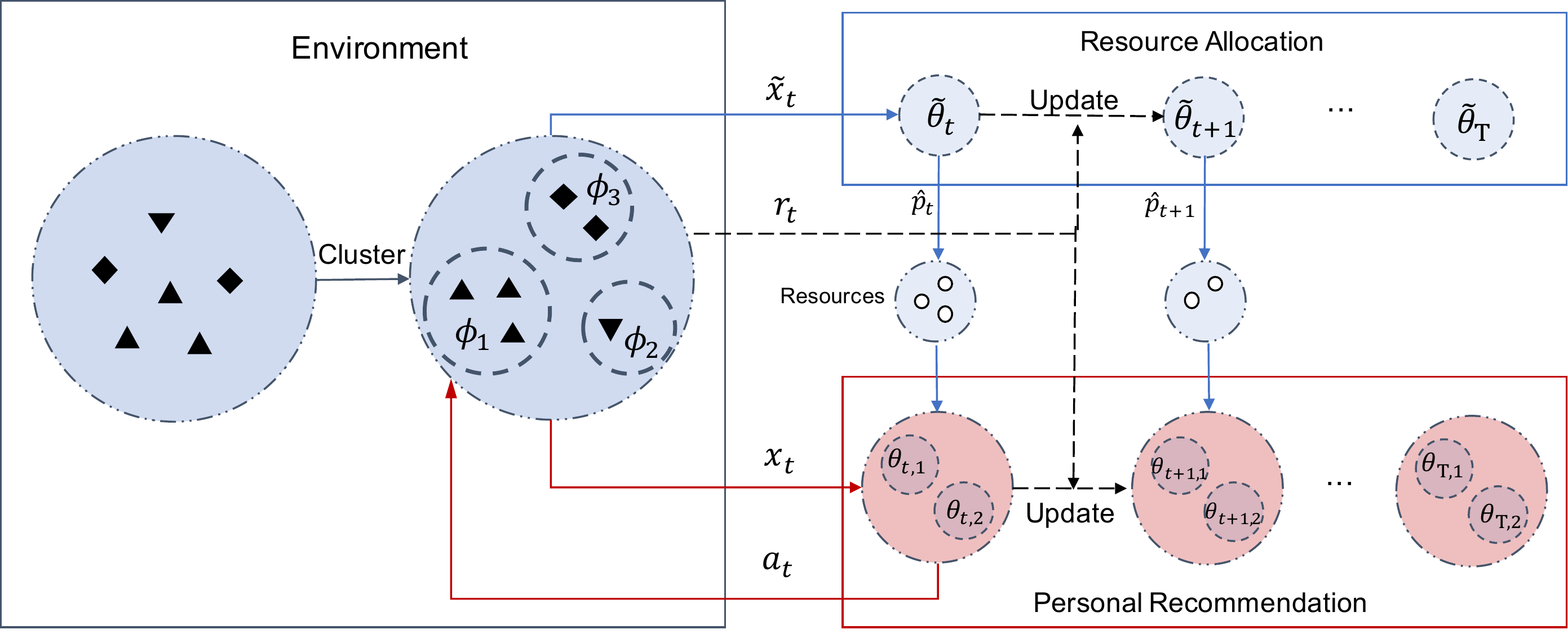}}
\caption{The framework of proposed HATCH: Environments provide historical user contextual features (include the mapping of user class distribution $\phi$ and user class center $\widetilde{X}$, which gained by a clustering method with massive historical user contextual data). In round $t$, firstly, the agent observe a user context ($x_t$), and get the class center ($\widetilde{x}_t$). Then, resource allocation level calculate the probability of retaining current context(retain current context with probably ($\hat{p}_t$)). Finally, personal recommendation level will choose an action according to the the estimation of the expect reward of each arms. After executing this action in the environment, the environment give a feedback ($r_{t}$) to the agent and update the parameters ($\widetilde{\theta},\theta$) in the algorithm.}
\label{fig1}
\end{figure}

\section{Related Work \& Preliminaries}
\subsection{Multi-armed Bandits}
Multi-armed bandits (MAB) problem is a typical sequential decision making process which is also treated as an online decision making problems. Bandit algorithm updates the parameters based on the feedback from the environment, and the cumulative regret measures the effect of policy learning. A number of non-linear algorithms have been proposed to solve this kind of problems \cite{filippi2010parametric}\cite{gopalan2014thompson}\cite{lu2018efficient}\cite{katariya2017bernoulli}\cite{yu2017thompson}. In additions, there are a lot of studies on the linear contextual bandits \cite{li2010contextual}\cite{abbasi2011improved}\cite{yue2012hierarchical}\cite{chou2014pseudo}.
MAB draws attention to a wide range of applications such as online recommendation system\cite{li2010contextual}\cite{pereira2019online}\cite{glowackabanditalgorithm}, online advertising\cite{chakrabarti2009mortal}\ and information retrieval\cite{glowacka2017bandit}. It provides solutions of exploration-exploitation trade-offs. Nowadays, MAB develop a lot of branches during the time, such as resource constraints bandits\cite{badanidiyuru2013bandits}, multi player bandits\cite{xia2016budgeted} and balanced bandits\cite{dimakopoulou2019balanced}. 

\subsection{Contextual Bandits}
Contextual bandits \cite{auer2002using,langford2007epoch} utilize the contextual feature information to make the choice of
the best arm to play in the current round. The extra feature information, known as 'context' are necessary in a lot of applications. To a large extent, it enhances the performance of bandits with relevant contextual information. In the generalized contextual multi-armed bandit problems, the agent observes a $d$-dimensional feature vector before making decision in round t. During the learning stage, the agent learns the relationship between contexts and rewards (payoffs). Contextual Thompson Sampling\cite{agrawal2013thompson}, LinUCB\cite{li2010contextual}, LinPRUCB \cite{chou2014pseudo}
analyzes the linear payoff functions. This paper is also based on the assumption of linear payoff function like LinUCB and focuses on the resource constraints situation. We briefly describe settings and formulation of LinUCB as follows: 

\begin{itemize}
\item We are in a k armed stochastic bandit system in each round $t$, the agent observes an action set $\mathcal{A}_t$ while the user feature context $x_t$ arrives independently.
%with identical distribution $\mathbb{P}\{x_t = j\} = {\phi}_j$. 

\item Based on observed payoffs in previous trials, the agent calculates the expectation of reward, which denotes as $r_{t,a_t}$ and the reward is modeled as a linear function $\mathbb{E}[r_{t}|x_{t,a_t}] = x^\top_{t,a_t}\theta_a^*$. After choosing an arm to play, it will receive the payoff cost $cost_{x_t,a_t}$. If the agent does not choose any arms, we denote it as $a_t = 0$, then $cost_{x_t,a_t} = 0$.

\item The agent chooses an arm $a_t\in \mathcal{A}_t$ by selecting the arm with maximum expected value at trial $t$.
\end{itemize}

\subsection{Bandit problems with resource constraints}
The resource-constrained multi-armed bandits are a family of bandits problem for conducting the policy learning in real-world scenarios, and choosing each arm to play will bring with the determined cost.
It is crucial to leverage the performance of policy learning as well as the available exploration resources. Previous works mainly focus on solving the problem in a finite state space of MAB setting with finite time horizon, including dynamic ads allocation with advertisers\cite{slivkins2013dynamic}, multi-play budgeted MAB\cite{xia2016budgeted}, thompson sampling to budgeted MAB \cite{xia2015thompson} and so on. \cite{agrawal2016linear,badanidiyuru2014resourceful} propose an algorithm which uses a knapsack method to address the problems in linear contextual bandits, it requires a prior distribution of contextual. \cite{wu2015algorithms} proposed a linear programming method to dynamically allocate the resource. However, this kind of method may not be suitable for infinite amount of contextual information in feature space.

\section{Method}
In this section, we present a hierarchical adaptive framework to balance the efficiency of policy learning and exploration of resources.

%For most bandits algorithms with resource constraint, they have not considered a situation that in some applications people are not willing to spend all their resources too early because the online environment is dynamic. Traditional resource constrain bandit algorithms always choose the action in greedy policy, which may cause the resource has been allocated too early(e.g., when put the trained policy to online environment, we can find that the resource are quickly run out). 

We first introduce the setting of budget constraint in the contextual bandits problem. Suppose that the total time-horizon is $\mathnormal{T}$ and total amount of resource $\mathnormal{B}$ is given, the total t-trial payoff is defined as $ \sum_{t = 1}^{T}r_{t,a_t}$ in the learning process. We denote the total optimal payoff as $\mathnormal{U}^*(T,B) = \mathbb{E}[\sum_{t = 1}^{T}r_{t,a_t^*}]$. Thus,  the objective is to maximize the total payoff during $\mathnormal{T}$ rounds under the constraints of exploration resource and time-horizon, we formulate the objective function as follows:
$$
 maximize\qquad \mathnormal{U}(T,B) = \mathbb{E}[\sum_{t = 1}^{T}r_{t,a_t}]
$$
$$
s.t.\qquad \sum_{t = 1}^{T}c_{x_t,a_t} \le B
$$
where $c_{x_t,a_t}$ is the associated cost when playing the arm $a_t$ for context $x_t$ at the $t$-th round. As a consequence, we denote the regret of the problem as follows:
\[R(T,B) = U^*(T,B) - U(T,B)\]
The objective function can be reformulated to minimize the regret $R(T,B)$.
In this paper, we propose a hierarchical structure to allocate the resource reasonably as well as to optimize the policy learning efficiently. In the upper level, our proposed HATCH adopts the adaptive resource allocation by considering the aspects of both users' information and remaining time/budget, balancing the payoff and gained reward distribution. In the lower level, HATCH utilizes the contextual information to predict the expectation of reward and make the decision to maximize the reward with the constraint of allocated exploration resource.

\subsection{Resource allocation globally}
It is challenging to directly allocate the  resources for exploration since we should not only consider the remaining time horizon but also the distribution of users' contextual information. It is a typical NP-hard problem because of infinite contextual state space and unknown reward distribution with time.  It is a challenge to conduct the policy learning since the feedback reward distribution is dynamic within the constraint of allowed  exploration resource. To simplify the problem, we divide the resource allocation process in two steps: the first one is to evaluate the user context distribution, we employ the massive historical logging dataset to map the user feature as state information into finite classes and evaluate contextual distribution. Then we dynamically allocate the resource by the expectation value of different user classes. We adopt an adaptive linear programming to solve the resource allocation problem. However, in real-world scenarios, the expectation of reward might not be obtained directly. Thus, we employ linear function to estimate the expectation of values. This alternative updating process can leverage the remaining exploration resource and obtained rewards.

\subsubsection{Dynamically resource allocation with expected reward.}
The exploration resource and time horizon might grow to infinite with the proportion of $\rho = B/T$, Linear Programming (LP) proved to be efficient to solve it \cite{wu2015algorithms}. When fixing the average resource constraint as $B/T$, LP function provides a policy whether choosing or skipping actions recommended by MAB. Further more, considering that remaining resource is changing constantly during the learning process, we replace resource $B$ with $b$ and time $T$ with time remaining $\tau$, which denote the remaining resource and remaining time-horizon in the $t$-th round. As a consequence, the averaged resource constraint can be replaced as $\rho = b_\tau/\tau$. We use a Dynamic Resource Allocation method (DRA) to address the dynamic average resource constraint. Different with \cite{wu2015algorithms}, the contexts' space in our study are infinite, which could not be presented numerically. 

We give some notation firstly. We denote $\mathcal{J}$ as finite user class which are clustered by the observed user contextual distribution. 
%Let $cost_{t,a_{t}}$ denote the occurred cost 
When executing the action $a_t$ in round $t$, there will be unit-costs of environment resource, which means that if the action is not dummy ($a_{t} \not= 0$) and is selected to execute, then the cost is equals to 1. In our recommendation framework, the actions will consume resources such as constraint number of display of the items. 
The quality of user class $j$ is captured by $u_{j}$ and is called expect reward of $j$. Then for every user classes, the
probability of occurrence is defined by a map $\phi_j(x)$ that means when the agent get an observation, it will find the class of this context and get probability of occurrence of this class by the distribution $\phi_j(x)$. In reality, the expected reward value are constants for each user class. We rank them by descending order: $u_1 > u_2 >...> u_J$, where $u_1$ is the largest one and $u_J$ is the smallest one among them. Unfortunately, we could not obtain the $u$ directly so that different with \cite{wu2015algorithms}, we define $\hat{u}_j$ as estimated reward for the class $j$ by linear function. In recommendation scenarios we focus on, there exists an assumption, that is

\textbf{Assumption 1} For any user class, in a time series $T$, the distribution $\phi_j(x)$ will not drift: $\phi_{j,t}(x) \sim \phi_{j,t+1}(x)$ for $t \in T$.

This assumption is intrinsic since user feature contexts $x$ are influenced only by user preference rather than the policy.

The objective of DRA is to make a decision whether the algorithm in the current round should retain the arm selection under the resource constraints or not. Let $p_j \in [0,1]$ be the probability that DRA retains the recommended arm at person level. We denote the probability vector as $\mathcal{P} = (p_1,p_2...,p_J)$. For given resource $B$ and time-horizon $T$, DRA can be formulated as follows:

\begin{equation}
({DRA}_{\tau,b})maximize \qquad\sum_{j = 1}^{J}p_j\phi_j u_j\nonumber
\label{eq2}
\end{equation}
\vspace{-2mm}
\begin{equation}\label{DRA}
s.t.\qquad \sum_{j = 1}^{J}p_j\phi_j\le \frac{B}{T}
\end{equation}
Let $\rho = \frac{B}{T}$ and $\tilde{j}(\rho) = max \{ j:\sum_{j^{'} = 1}^{j} \phi_{j^{'}} \le \rho\}$, which represents a threshold of averaged budget.

Like the formulation in \cite{wu2015algorithms}, the optimal solution of DRA can be summarized as following:
$$
p_j(\rho) = \left\{
\begin{aligned}
&1,& & \textbf{if}& & 1\le j \le\widetilde{j}(\rho)\\
&\frac{\rho-\sum_{j' = 1}^{\widetilde{j}(\rho)}{\phi_{j'}}}{\phi_{\widetilde{j}(\rho)+1}},& & \textbf{if}& & j = \widetilde{j}(\rho)+1\\
&0,& & \textbf{if}& & j>\widetilde{j}(\rho)+1
\end{aligned}
\right.
$$
We denote $p_j(\rho)$ as the solution of eq.\ref{DRA} and $v(\rho)$ as the maximum expected reward in a single round within averaged resource. However, in real-world scenarios, it can not guarantee a static ratio of resource and time horizon. Thus, we replace the static ratio $\rho$ as $b_{\tau}/\tau$, where $b_{\tau}$ and $\tau$ represent the remaining resources and time in round $t$. 

\subsubsection{Estimate expected rewards of user classes.} 

To solve the problem that $u_j$ is difficult to obtain in real-world scenarios, we should estimate the expectation of $u_j$. Firstly, we map the observed user context into finite user classes. For each user class, there is a representation center point which is denoted by $\widetilde{x}$ and $j$ is the $j$-th cluster. When observing a context $x_t$ in round $t$, it will automatically map to $\widetilde{x}_t$ and we estimate the value ($\hat{u}_{t,j}$) for this observation.

We adopt a linear function to estimate the expectation reward: $\mathbb{E}[u|\widetilde{x}] = \widetilde{x}^\top \widetilde{\theta}_j$ between the contextual information $\widetilde{x}$ and the reward $r$. We normalize the parameters with $\|x\|\le1$ and $\|\widetilde{\theta}\|\le1$.

We set the matrix $\widetilde{X}_j = [\widetilde{x}_1,\widetilde{x}_1...\widetilde{x}_t]$ to denote all the historical observations of the user class $j$, where $\|\widetilde{x}\| \le 1$ and every vector in $\widetilde{X}_j$ is equal to $\widetilde{x}_j$. 

When formulating reward evaluate function of each user class as a ridge regression, we can get the following equation:
\begin{equation}\label{tildetheta}
\widetilde{\theta}_{t,j} = \widetilde{A}_{t,j}^{-1}\widetilde{X}_{t,j}Y_{t,j}^\top
\end{equation}
where $Y_{t,j} = [r_1, r_2 .. r_t]$ and $\widetilde{A}_{t,j} = (I + \widetilde{X}_{t,j}^\top \widetilde{X}_{t,j} )$. 

Let $\|\widetilde{x}_{t}\|_{\widetilde{A}^{-1}} = \|\widetilde{s}_{t}\| = \|\sqrt{\widetilde{x}_{j}^\top \widetilde{A}_{t,j}^{-1} \widetilde{x}_{j}}\|$, and $\widetilde{\theta}_{j}^*$ denotes the expect value of $\widetilde{\theta}_{j}$ and $u_j =  \widetilde{x}_{j}^\top \theta^*_{j} $. The regret of each round can bounded by a confidence interval:
$$
|\widetilde{x}_{j,t}(\widetilde{\theta}_{j,t} - \widetilde{\theta}^*_j)|\le(1+\widetilde{\alpha}) \widetilde{s}_{t}
$$
where $\widetilde{\alpha} = \sqrt{\frac{log(2/\delta)}{2}}$.

We use the estimated expectation of reward $\hat{u}_{t,j} = \widetilde{x}_{j}^\top \widetilde{\theta}_{t,j}$ to solve DRA and get the $\hat{p}_x$ to allocate the resource. 

The following property for estimation processing is crucial to guarantee regret bound of proposed HATCH. When the number of times to execute the algorithm is larger than a constant, the algorithm can get the correct order of $u$.

\textbf{Lemma 1} For two user class pairs $(j,j{'})$, let  $N_j(t-1)$ be the number that $j$-th class appears until round $t-1$. If the expectation of reward satisfies $u_j<u_{j'}$, then for any $t\le T$,

\begin{equation}\label{eq6}
\mathbb{P}( \hat{u}_{j,t} \ge \hat{u}_{j',t}| N_j(t-1)\ge l_j) \le 2t^{-1}
\end{equation}
where $l = \frac{2logT}{(u_j-u_{j'})^2}$. Most of proofs are shown in appendix section.

Lemma $1$ illustrates that when the algorithm executed for $N$ times, the order of estimation of reward expectation is the same as the actual order of it.

\begin{algorithm}[t]
\caption{Hierarchical AdapTive Conextual bandit metHod (HATCH)}
\begin{algorithmic}[1]
\REQUIRE $\lambda$,$B$, $\mathcal{A}$, $\alpha$, $\widetilde{\alpha}$
\STATE Init $\tau = T$, $b = B$, $\hat{u}_{0,j} = 1$

\STATE Map the historical context into finite user class set $\mathcal{J}$, obtain $\phi$ for each user class distribution.
\STATE Init $\widetilde{A}_{0,j} = \textbf{I}$, $\widetilde{\theta}_{0,j} = \textbf{0}$,$\widetilde{X}_{0,j} = \emptyset$, $Y_{0,j} = \emptyset$, $\forall j\in\mathcal{J}$
\STATE Init $A_{0,j,a} = \textbf{I}$, $\theta_{0,j,a} = \textbf{0}$, $X_{0,j,a} = \emptyset$, $Y_{0,j,a} = \emptyset$, $\forall j\in\mathcal{J}$  and  $\forall a \in \mathcal{A}$

\FOR{ $t = 1,2,...T $}
    \STATE Observe the context information $x_t$, get the context class $j$ of $x_t$, and obtain the mapped user class context  $\widetilde{x}_{t}$.
    \STATE Get action $a$ by calculate the eq.\ref{selecta}
    \IF{$b >0$}
    \STATE Obtain the probabilities $\hat{p}_j(b/\tau)$ by solving $DRA{(\tau,b)}$ and with ${u}$ replaced by $\hat{u}$.
    \STATE Take action $a$ with probability $\hat{p}_j(b/\tau)$
    \ENDIF
    \STATE Observe reward $r_{t,a}$ from the environment.
    \STATE Update $\tau, b$
    \STATE Define $X_{t,j,a} \leftarrow
 [X_{t-1,j,a}:x_{t}]$
    \STATE Define $Y_{t,j,a} \leftarrow [Y_{t-1,j,a}:r_{t,a}]$
    \STATE Define $\widetilde{X}_{t,j} \leftarrow [\widetilde{X}_{t-1,j}, \widetilde{x}_{t}]$
    \STATE Define $Y_{t,j} \leftarrow [Y_{t-1},r_{t,a}]$
    \STATE $\widetilde{A}_{t,j} \leftarrow I + \widetilde{X}_{t,j}^\top\widetilde{X}_{t,j}$ 
    \STATE $\widetilde{\theta}_{t} \leftarrow \widetilde{A}_{t,j}^{-1}\widetilde{X}_{t,j}Y_{j,t}$
    \STATE $\hat{u}_{t,j} \leftarrow \widetilde{x}_t^\top\widetilde{\theta}_{t,j}$
    \STATE ${A}_{t,j,a} \leftarrow \lambda I + {X}_{t,j,a}^\top{X}_{t,j,a}$
    \STATE ${\theta}_{t,j,a} \leftarrow {A}_{t,j,a}^{-1}{X}_{t,j,a}Y_{t,j,a}$

\ENDFOR
\end{algorithmic}
\end{algorithm}
\subsection{Personalized recommendation level}

In the personalized recommendation and policy learning level, we utilize the full user contextual information $x_t$ to conduct the policy learning and recommend the best action to play. Following the setting of contextual bandit from \cite{abbasi2011improved}, we use a linear function to fit the model of context and reward: $\mathbb{E}[r|x_{t}] = x_{t}^\top\theta_{t,j,a}$.

We set the personal information context matrix as
$$
X_{t,j,a} = {[x_1, x_2, ... x_t]}
$$
where $x_t$ is in user class $j$ and choose the action $a$.

Similar with eq.\ref{tildetheta}, we have the policy coefficient vector in the $j$-th class with $a$-th action:
$$ \theta_{t,j,a} = {A}_{t,j,a}^{-1}X_{t,j,a}Y_{t,j,a}$$
where ${A}_{t,j,a} = (\lambda I + {X}_{t,j,a}^\top {X}_{t,j,a} )$.

We set $r =  x_{t}^\top\theta^*_{j,a} + \epsilon$, supposing that $\epsilon$ is 1-sub-gaussian independent zero-mean random variable, where $\mathbb{E}[\epsilon] = 0$ and $\theta_{j,a}^*$ denotes the expect value of $\theta$.

We choose the arm which maximum reward expectation in the action set $\mathcal{A}$ :
\begin{equation}\label{selecta}
    a_t^* = argmax_{a\in\mathcal{A}}x_t^\top\theta_{t,j,a} + (\sqrt{\lambda}+\alpha)\|x_{t}\|_{A^{-1}}
\end{equation}
$$
\alpha = \sqrt{2log(\frac{det(A_{t,j,a})^{\frac{1}{2}},det(\lambda I)^{\frac{1}{2}}}{\delta})}
$$
where $\delta$ is a hyperparameter and $\lambda>0$ is a regularized parameter.

Then, we use the $p_j$ calculated by DRA to decide whether to execute the action $a_t^*$ or not. The whole process of the proposed HATCH method is illustrated in Algorithm $1$ and the algorithm pipeline is shown in Figure 1.

\section{Regret Analyze of Hierarchical Adaptive Contextual Bandits}

In this section, we present a theoretical analysis on the regret bound of our proposed HATCH.

In round $t$, the upper bound is summarized as follows:
\begin{align}
v_{t}(\rho) = \sum_{j = 1}^{j( \rho)}\phi_ju_{j,t}^* + p_{\widetilde{j}(\rho)+1}(\rho) \phi_{\widetilde{j}(\rho)+1} u_{{j,t}_{\widetilde{j}(\rho)+1}}^*\nonumber
\end{align}
where $u_{j,t}^* = max_{a\in\mathcal{A}}x_{t,j,a}^\top\theta^*_{j,a}$ denotes the optimal reward for personalized recommendation level in round $t$.
The regret of our algorithm is:
\begin{equation}\label{eq3}
R(T,B) = U^*(T,B) - U(T,B)\nonumber
\end{equation}
\textbf{Theorem 1}
Given $\phi_j$, $u_j$ and a fixed $\rho\in(0,1)$, let $\Delta_j = inf\{|{u}_{j'}-{u}_j|\}$, where $j' \in J$ and $j' \not= j$. Then let $q_j=\sum_{j' = 1}^{j}\phi_{j'}$. For any $j\in \{1, 2,...J\}$, the regret of HATCH satisfies:

1)\textbf{(Non-boundary cases)} if $\rho \not= q_j$ for any $j\in\{1,2...J\}$, then 
$$
R(T,B) = O(J\beta_j\sqrt{\Phi logT log(\Phi logT)}+\beta\sqrt{Blog(B)}+JlogT)
$$

2)\textbf{(Boundary cases)} if $\rho = q_j$ for any $j\in\{1,2...J\}$, then 
$$R(T,B) = O(\sqrt{T} + J\beta_j\sqrt{\Phi logT log(\Phi logT)}+\beta\sqrt{Blog(B)}+JlogT)$$
where $\Phi = \frac{1}{\Delta^2}+1$,

$$\beta_j = \sqrt{\lambda}+\sqrt{2log(1/\delta) + log(1+2(logT/\Delta_j^2+logT+1)/\lambda))}$$
$$\beta = \sqrt{\lambda}+\sqrt{2log(1/\delta) + log(1+B/\lambda))}$$

The sketch of proofing include three step. 

\textsl{Step 1: Partition of regret}
. We divide the regret into two parts. $\mathbb{E}[N_j(T)]$ denotes the expected number of times that the $j$-th user class is allocated resource when order error occur. And $N_j$ is the total number that class $j$ is allocated resources until round $T$:
% & \sum_{t = 0}^{T}v_{t,\rho}(u^*-\widetilde{u}^*) +\\
\begin{align}
    &  R(T,B)= U^*(T,B) - U(T,B)\nonumber\\
    % =& \sum_{t = 0}^{T}v_{t}(\rho)- \sum_{j = 1}^{J}u_j\mathbb{E}[N_j(T)]\nonumber\\
    \le& [\sum_{t = 0}^{T}v_{t}(\rho)- \sum_{j = 1}^{J}{u}_j^*\mathbb{E}[N_j(T)]] +
    %  & [\sum_{t = 1}^{T}\sum_{j = 1}^{J}\sum_{a = 1}^{\mathcal{A}}x_{t}({\theta}_j^*-{\theta}_{j,a})]\nonumber\\
      [\sum_{j = 1}^{J}\sum_{t = 1}^{N_j}\sum_{a = 1}^{\mathcal{A}}x_{t}({\theta}_{j,a}^*-{\theta}_{j,a})]\nonumber\\
    =&  R^{(2)}(T,B) +R^{(1)}(T,B)\label{eq1}
\end{align}
where $u_j^* = u_{j,a^*} = u_j + \eta^*$, $\eta^*$ is a random variable which measures the difference of expected reward between user class $j$ and optimal selected action $a^*$ for $x_{t,j}$.

\textsl{Step 2: Bound of $R^{(1)}(T,B)$:}

For $j \not= j'$, let  $\Delta_j = inf\{|{u}_{j'}-{u}_j|\}$, where $j' \in J$.
% \begin{align}
%     &\mathbb{E}[N_j(T)]\nonumber\\ \le
%     &l_{j}+\sum\mathbb{P}\{\widetilde{x}_{t,j},N_j(T)\ge l_{j}\}\nonumber\\ \le
%     &l_{j}+\sum_{t = 1}^{T}2t^{-1}
% \end{align}
% Following the facts that $\sum_{t = 1}^{T}t^{-1}\le1+\log T$. Then according to Lemma $2$, $R^{(1)}(T,B)$ can be bounded as 
\begin{align}
    R^{(1)}(T,B)
    % [\sum_{j = 1}^{J}\sum_{a = 1}^{\mathcal{A}}x_{t}({\theta}_j^*-{\theta}_{j,a})]\nonumber\\
    % &\le\sum_{j = 1}^{J}\mathbb{E}[N_j(T)]\sum_{a=1}^{\mathcal{A}}|x_t(\hat{\theta}_{j,a}-\theta_{j,a}^*)|\nonumber\\
    % % &\le \sum_{j = 1}^{J}\mathbb{E}[N_j(T)]\sum_{a = 1}^{\mathcal{A}}\|x_t\|_{A^{-1}}(\sqrt{2log(\frac{det(A)^{\frac{1}{2}}}{\delta})}+\|\theta^*\|)\nonumber\\
    % & \le \sum_{j = 1}^{J}\mathbb{E}[N_j(T)]\sum_{a = 1}^{\mathcal{A}}\|x_t\|_{A^{-1}}(\alpha + 1)\nonumber\\
    &\le \sum_{j = 1}^{J}\beta_j(\sqrt{\mathbb{E}[N_j(T)]log(\lambda+\mathbb{E}[N_j(T)])})\nonumber\\
    &+\beta\sqrt{{B}log(\lambda+B)}\label{equ:2}
\end{align}
where 
$$
\mathbb{E}[N_j(T)] = \frac{2logT}{\Delta_j^2}+2logT +2,
$$
$$\beta_j = \sqrt{\lambda}+\sqrt{2log(1/\delta) + log(1+2(logT/\Delta_j^2+logT+1)/\lambda))}$$
$$\beta = \sqrt{\lambda}+\sqrt{2log(1/\delta) + log(1+B/\lambda))}$$
Most part of proofs are shown in appendix section.

\textsl{Step 3: Bound of $R^{(2)}(T,B)$:}

% In UCB-ALP setting, each context has totally A arm, the update process is like Bernoulli trial and they deal with the linear programming by $u_j^* = max_{k} u_{j,a}$. And

In our method, actually, the estimation of resource allocation level can be treated as a Bernoulli trial. For personalized recommendation level,  the recommendation is a linear contextual bandits trial. 
% In these two settings, 
Each arm in each user class has optimal expect reward. 
% the difference between HATCH and ALP-UCB is the strategy of choosing arm - 
In HATCH, we use linear contextual bandits to evaluate expected rewards and the intrinsic user preference ($\theta^*$), which defined by arms' property in each user class.
% , but also the observed context detail in each turn. 
% Thus $u^*_j$ in our setting is the same as $u^*$ in UCB-ALP. 
It means that $\eta^*$ are known constant which is lower than a positive number, $\eta^* \le C$.

In this study, $R^{(2)}(T,B)$ is only associated with global level user classes since we use the estimation of $u_j$ to solve DRA and get $\hat{p}$. Thus, all the errors in this part come from the order error of $\hat{u}$. The error includes three events: the first one is roughly correct order event during the learning process, the other two are incorrect classes order situations which are defined as follows\cite{wu2015algorithms}:
\begin{align}
\varepsilon_0(t) = &\{\forall j \le \widetilde{j}(\rho), \hat{u}_j(t)>\hat{u}_{\widetilde{j}(\rho)+1}(t);\nonumber\\
&\forall j> \widetilde{j}(\rho)+1,\hat{u}_j(t)< \hat{u}_{\widetilde{j}(\rho)+1}(t)\}\label{equ:3}
\end{align}
\begin{align}
\varepsilon_1(t) = &\{\exists j \le \widetilde{j}(\rho), \hat{u}_j(t)\le\hat{u}_{\widetilde{j}(\rho)+1}(t);\nonumber\\
&\forall j> \widetilde{j}(\rho)+1,\hat{u}_j(t)< \hat{u}_{\widetilde{j}(\rho)+1}(t)\}\label{equ:4}
\end{align}
\begin{align}
\varepsilon_2(t) = &\{\exists j > \widetilde{j}(\rho)+1, \hat{u}_j(t)\ge\hat{u}_{\widetilde{j}(\rho)+1}(t)\}\label{equ:5}
\end{align}
Define that $v^*(\tau,b_{\tau}) = \sum_{j = 1}^{J} \widetilde{p}_j(b_{\tau}/\tau)\phi_ju_{j}^*$, where $u_{j}^* = u_j+\eta^*$. The single-round difference between the algorithm and upper bound $v(\rho)$ is
$$
\Delta v_t = v(\rho) - v^*(\tau, b_{\tau})
$$
Then we have $ R^{(2)}(T,B) = \sum_{\tau = 1}^T\mathbb{E}[\Delta v_{\tau}]$. Considering all the possible situations, the expectation can be rewritten as
\begin{equation}
    \mathbb{E}[\Delta v_{\tau}] = \sum_{s = 0}^2\mathbb{E}[\Delta v_{\tau}, \varepsilon_s(T-\tau+1)]
\end{equation}
Let $\bar{u} = \sum_{j = 0}^{J}u_j$. We can get the upper bound of $R^{(2)}(T,B)$ in our setting in two cases:

\textbf{Non-boundary cases:}
\begin{align}
    &\limsup_{t\rightarrow\infty}\frac{R^{(2)}(T,B)}{logT}\nonumber\\
\le   &[\bar{u} + C + v(\rho)]
[\sum_{j = 1}^{\widetilde{j}(\rho)}\frac{27}{2g_{\widetilde{j}(\rho)+1}[\Delta_{\widetilde{j}(\rho)+1}]^2}\nonumber\\
+ &\sum_{j = \widetilde{j}(\rho)+2}^{J}\frac{27}{2g_j[\Delta_{j}]^2}+ 2J]\label{eq4}
\end{align}

where $g_j = min \{\phi_j, \frac{1}{2}(\rho-{q_{\widetilde{j}(\rho)}}), \frac{1}{2}(q_{\widetilde{j}(\rho)+1}-\rho) \}$

\textbf{Boundary cases:}

Let $R^{(2)}_1(T,B) = \sum_{\tau = T}^{1}\mathbb{E}[\Delta v_{\tau}, \varepsilon_0(T-\tau+1)$, and $R^{(2)}_2(T,B) = \sum_{\tau = T}^{1}\sum_{s = 1}^2\mathbb{E}[\Delta v_{\tau}, \varepsilon_s(T-\tau+1)$
\begin{align}
    &R^{(2)}_1(T,B)\le  (u_1+C)\sqrt{\frac{Var(b_\tau)}{\tau^2}}\ + 2v(\rho)e^{-2\zeta^2\tau}\label{eq5}
\end{align}
\begin{align}
    &\limsup_{t\rightarrow\infty}\frac{R^{(2)}_2(T,B)}{logT}\nonumber\\
\le   &[\bar{u} + C + v(\rho)]
[\sum_{j = 1}^{\widetilde{j}(\rho)}\frac{27}{2g_{\widetilde{j}(\rho)+1}[\Delta_{\widetilde{j}(\rho)+1}]^2}\nonumber\\
+ &\sum_{j = \widetilde{j}(\rho)+1}^{J}\frac{27}{2g_j[\Delta_{j}]^2}+ 2J]\label{eq4}
\end{align}

where $\zeta = \frac{1}{2}min\{\rho-q_{\widetilde{j}(\rho)-1}, q_{\widetilde{j}(\rho)+1}-\rho\}$
$$g_j = min\{\phi_j, \frac{1}{2}(\rho-{q_{\widetilde{j}(\rho)-1}}), \frac{1}{2}(q_{\widetilde{j}(\rho)+1}-\rho)\}$$

Most part of proof are shown in appendix section.

\begin{figure*}
\centering
\subfigure[]{
\begin{minipage}[t]{0.32\linewidth}
\centering
\includegraphics[width=1\textwidth]{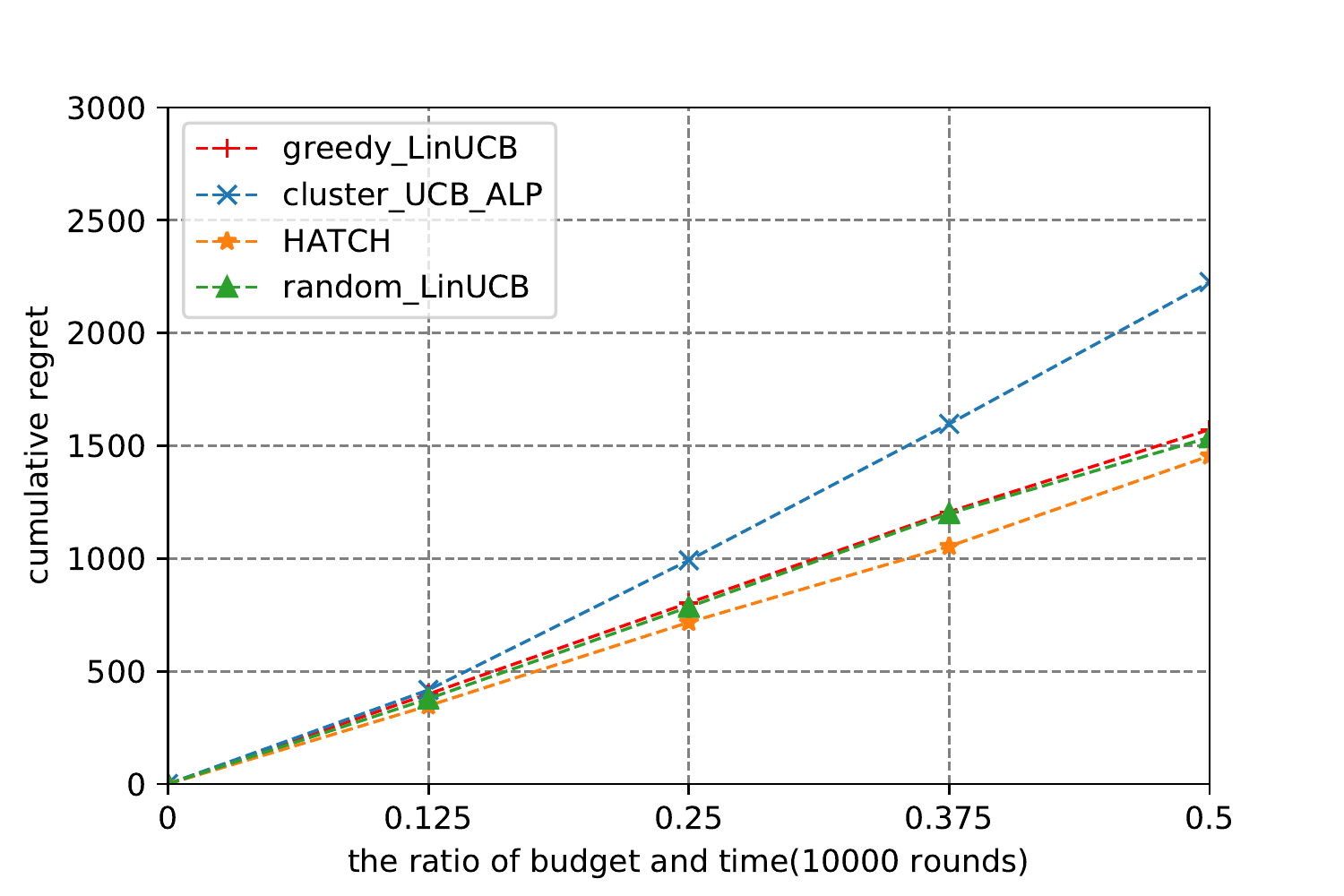}
%\caption{fig1}
\end{minipage}%
}
\subfigure[]{
\begin{minipage}[t]{0.32\linewidth}
\centering
\includegraphics[width=1\textwidth]{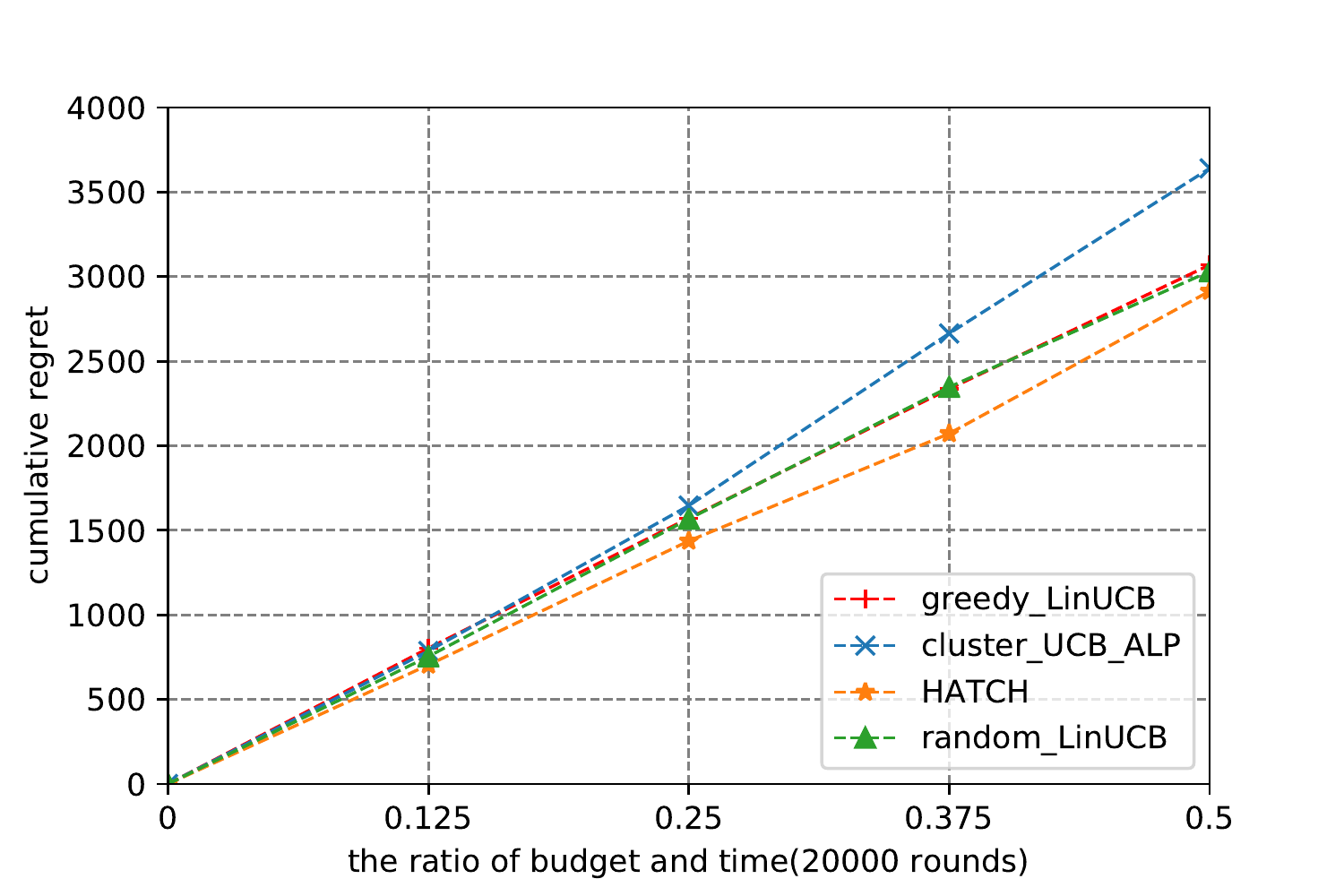}
%\caption{fig1}
\end{minipage}%
}
\subfigure[]{
\begin{minipage}[t]{0.32\linewidth}
\centering
\includegraphics[width=1\textwidth]{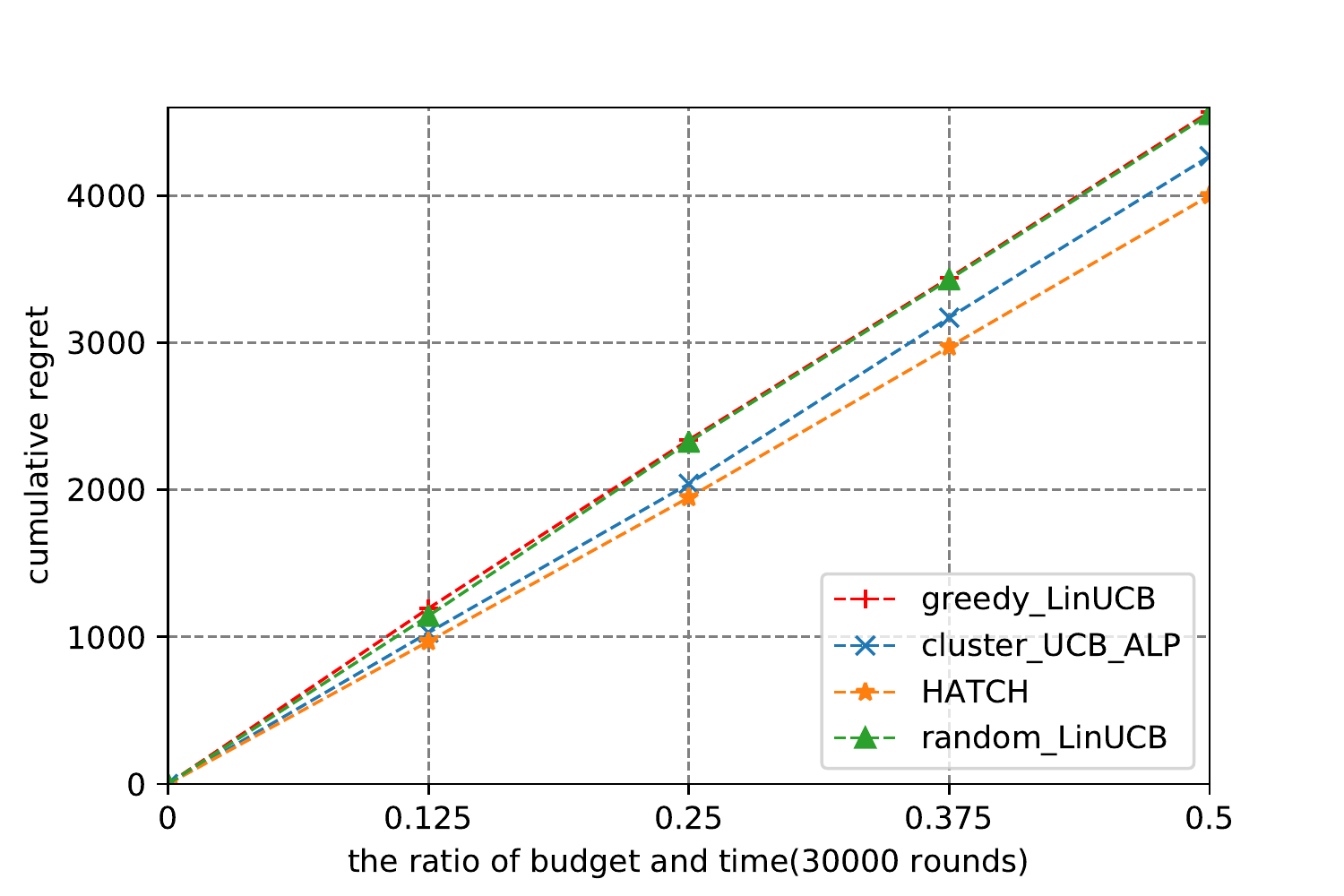}
%\caption{fig1}
\end{minipage}%
}
% \subfigure[]{
% \begin{minipage}[t]{0.24\linewidth}
% \centering
% \includegraphics[width=1\textwidth]{avarage_reward.pdf}
% %\caption{fig1}
% \end{minipage}%
% }
\caption{ This figure shows the synthetic evaluation of HATCH as well as compared methods, with the cumulative regret for different ratio of $\rho = \{0, 0.125, 0.25, 0.375, 0.5\}$. (a), (b), (c) illustrate different execution rounds.}
\label{fig5}
\end{figure*}

\section{Experiments}
In this section, we conduct two groups of experiments to verify the effectiveness of the proposed method. The first one is a synthetic experiment %in which we show the regrets and resource usage by comparing to state-of-the-art methods, 
and the other is conducted on the Yahoo! Today data set. The implementation details are available on the GitHub. \footnote{https://github.com/ymy4323460/HATCH}

\subsection{Synthetic evaluation}
For synthetic evaluation, we build a synthetic data generator. We set the dimension of each generated context is $dim = 5$ and the value range of each dimension is in $[0,1]$. We set $J = 10$ as the number of user classes and $K = 10$ arms with the reward generated from a normal distribution. Then we set the distribution of user class is $[0.025, 0.05, 0.075, 0.15, 0.2, 0.2, 0.15, 0.075, 0.05, 0.025]$ and the expected reward $u_j$ of each user class is a random number in $[0,1]$. Each arm has its random expected reward number $\sigma_{j,a}$ which is the sum of user class expected reward and arm expected reward ($u_{j,a} = u_j + \sigma_{j,a}$). Every dimension in weights $w_{j,a}$ are random in $[0,1]$ and $\|w_{j,a}\| \le 1$. We sampled 30000 synthetic contexts from data generator with the distribution $\phi$ and we denote $X_j $ as context set of class $j$. Rewards of a context have 10 number corresponding to 10 arms and they are generated from normal distributions with the mean of $u_{j,a}+x_{j}w_{j,a}$ and variance of $1$. Rewards are normalized  be 0 or 1. 

Experiments are conducted by comparing with state-of-the-art algorithms. The methods and setting are described as follows:

\textbf{greedy-LinUCB} \cite{li2010contextual} adopts the LinUCB strategy and chooses the best arm in each turn when the choice is executed, consuming one unit of resource. This process will keep running until the resource is exhausted.

\textbf{random-LinUCB} is the LinUCB algorithm which choose the best arm in each turn. The choice is executed with the probability of $\rho = (b_\tau/\tau)$. The resource allocation strategy is straightforward but it reduces the probability to miss high valuable contexts in the long time horizon.

\textbf{cluster-UCB-ALP} \cite{wu2015algorithms} proposes an adaptive dynamic linear programming method for UCB problems. For synthetic evaluation, we use the index of user class as the input at round $t$. This method only counts the reward and the number of occurrences for each user class and will not use class features due to the UCB setting.

Since the regrets of all the methods are not identical, we compare the accumulate regret (the optimal reward minus the reward of executed actions) when the choices are executed. We set four different scenarios with time and budget constraints: $\rho = \{0, 0.125, 0.25, \\ 0.375, 0.5\}$ with different execution rounds (10000, 20000, 30000). 

In Figure 2, we can observe that HATCH achieves the lowest cumulative regrets among all the methods. In addition, when the budget-time ratio $\rho$ is gets larger, the regrets become larger.

In all conditions, the regret of HATCH in Figure 2 is lower than random-LinUCB which indicates our algorithm retain the high valuable user contexts' choice. The charts show that our algorithm performs better than cluster UCB-ALP, because context have lots of information that UCB setting can not utilize and the regret of UCB-ALP is higher than other linear contextual methods. The regret of random-LinUCB is close to greedy-LinUCB.  The difference between greedy-LinUCB and random-LinUCB is that greedy-LinUCB spent all the resource in early time and the resource usage is more uniform in random-LinUCB. For the synthetic data, the reward function is nearly static during all learning rounds. Thus, learning in the early time (greedy-LinUCB) seems to be the same as random resource allocation strategy. However, in real-time setting, this distribution might not be stable during the learning process. 

%One interesting phenomenon is that at the early time cluster-LinUCB has the largest regrets since the estimation of $u_{j,a}$ is not stable, and by using this value to solve the DRA problem will lead the unstable of $\hat{p}$. This algorithm maybe need large training process to get warm start then it can perform better.

\subsection{Experiments on Yahoo! Today Module}
The second group of experiments are conducted on the real-world scenarios which involve news article recommendation in the “Today Module”, a common benchmark to verify performance of recommendation algorithms. Data are collected from on the Yahoo! front page \footnote{https://webscope.sandbox.yahoo.com/catalog.php?datatype=r\&did=49}.  When users visit this module, it will display high-quality news articles from candidate articles list. The data set are collected at Yahoo! front page for two days at May 2009. It contains T = 4.68M events in the form of triples $(x, a, r)$, where the context $x$ contains user/article features. The user features have a dimension of $6$ while $a$ denotes the candidate article with the reward $r$ of the click signal (1 as click and 0 not click). We random select T = 1.28M events with the top 6 recommended articles and fully shuffled these data before conducting the learning process. 

We use half of contextual feature dataset to obtain a pre-defined mapping method. Here we choose Gaussian Mixture Model as mapping method in our system, denoted by $\mathcal{G}(x)$. We set $J = 10$ cluster centers (The number of contextual centers is suggested to be larger than the feature dimension). The distribution of all clusters is denoted by $\phi$ which provided by GMM.  

\textbf{Evaluator}: we evaluate the algorithm in the offline setting. We use the strategy of reject sampling from the historical data as \cite{li2010contextual}. However, this kind of samples and evaluate strategy is not suitable for our setting. In our setting, the user classes must have a determined distribution during the learning process. In the setting of rejected sample evaluation in \cite{li2010contextual}, the distribution of context class sometimes are not stable at the early time and it will lead to the choice of arms concentrates on several arms. For example, if algorithm chooses the best arm $a$ and this user context-arm pair did not occur in historical data, this context will be thrown away, causing the user classes distribution to drift at the early time and it leads to the large variance of experiments result. Thus, we propose a new evaluation function to select data in a static user class distribution. 

We show the evaluation strategy in Algorithm 2, adjusting reject sampling evaluation proposed by \cite{li2010contextual} to fit our setting.

\begin{figure*}
\centering
\subfigure[$\rho = 0.125$]{
\begin{minipage}[t]{0.225\linewidth}
\centering
\includegraphics[width=1\textwidth]{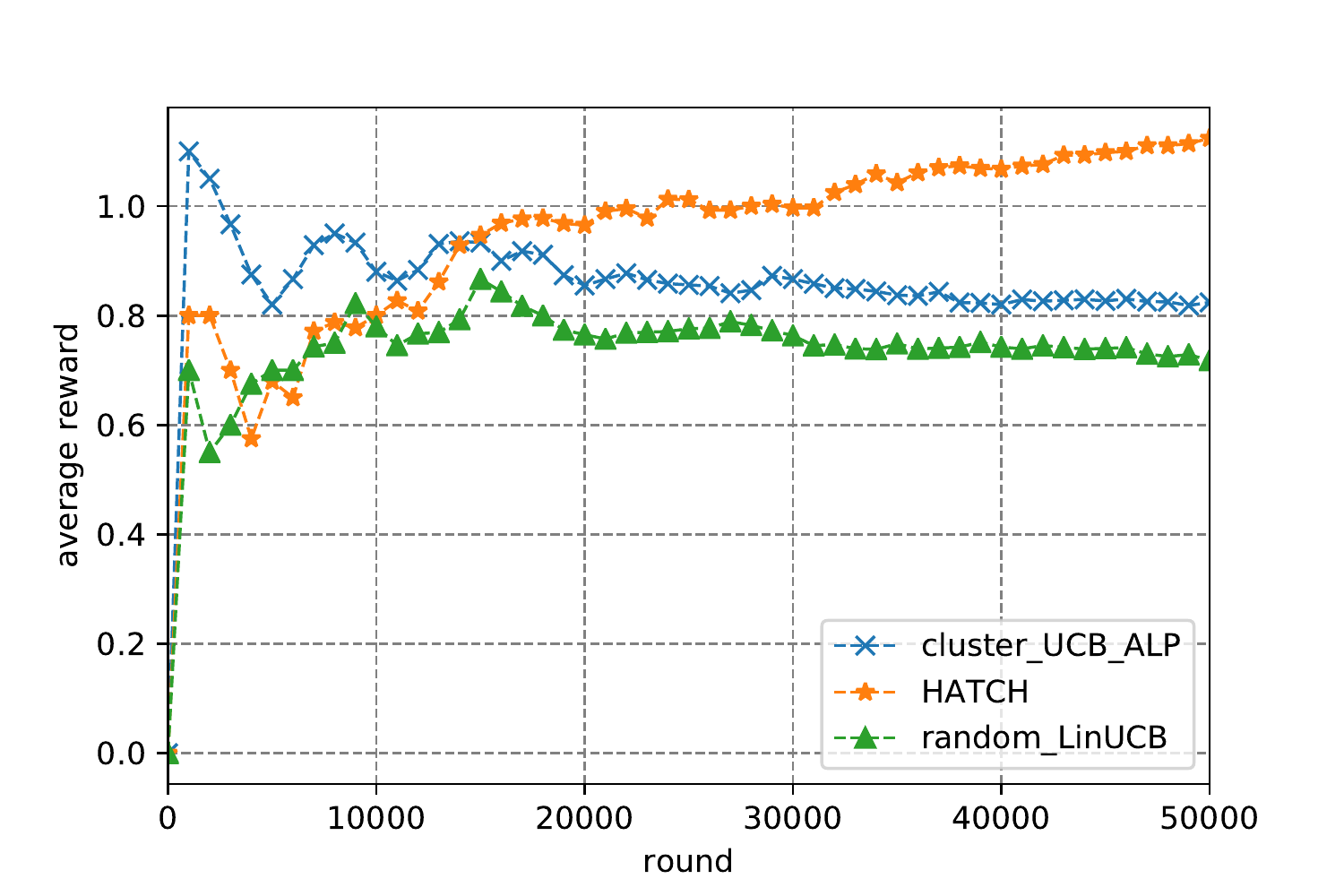}
% \caption{$\rho = 0.125$}
\end{minipage}%
}
\subfigure[$\rho = 0.25$]{
\begin{minipage}[t]{0.225\linewidth}
\centering
\includegraphics[width=1\textwidth]{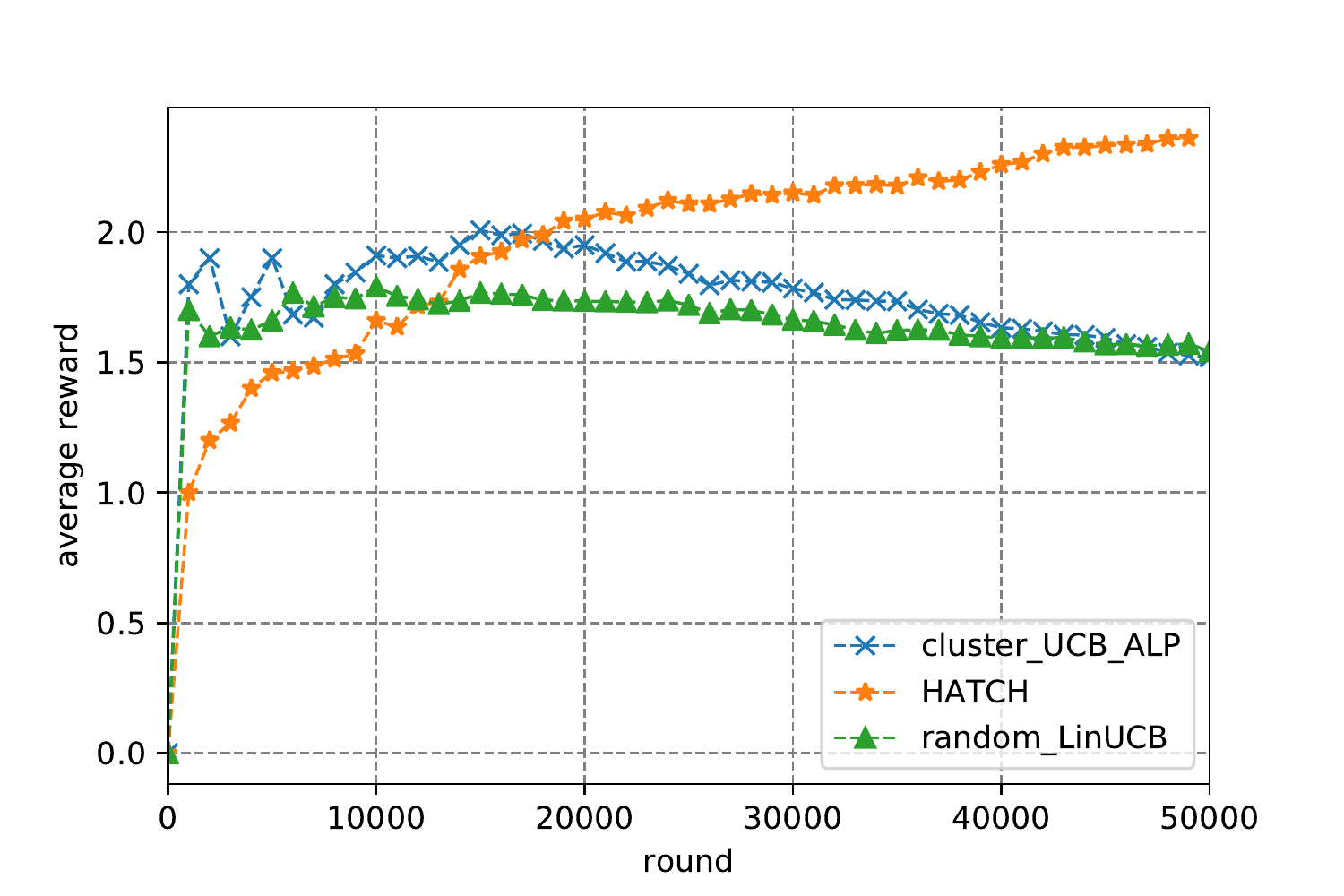}
% \caption{$\rho = 0.25$}
\end{minipage}%
}
\subfigure[$\rho = 0.375$]{
\begin{minipage}[t]{0.225\linewidth}
\centering
\includegraphics[width=1\textwidth]{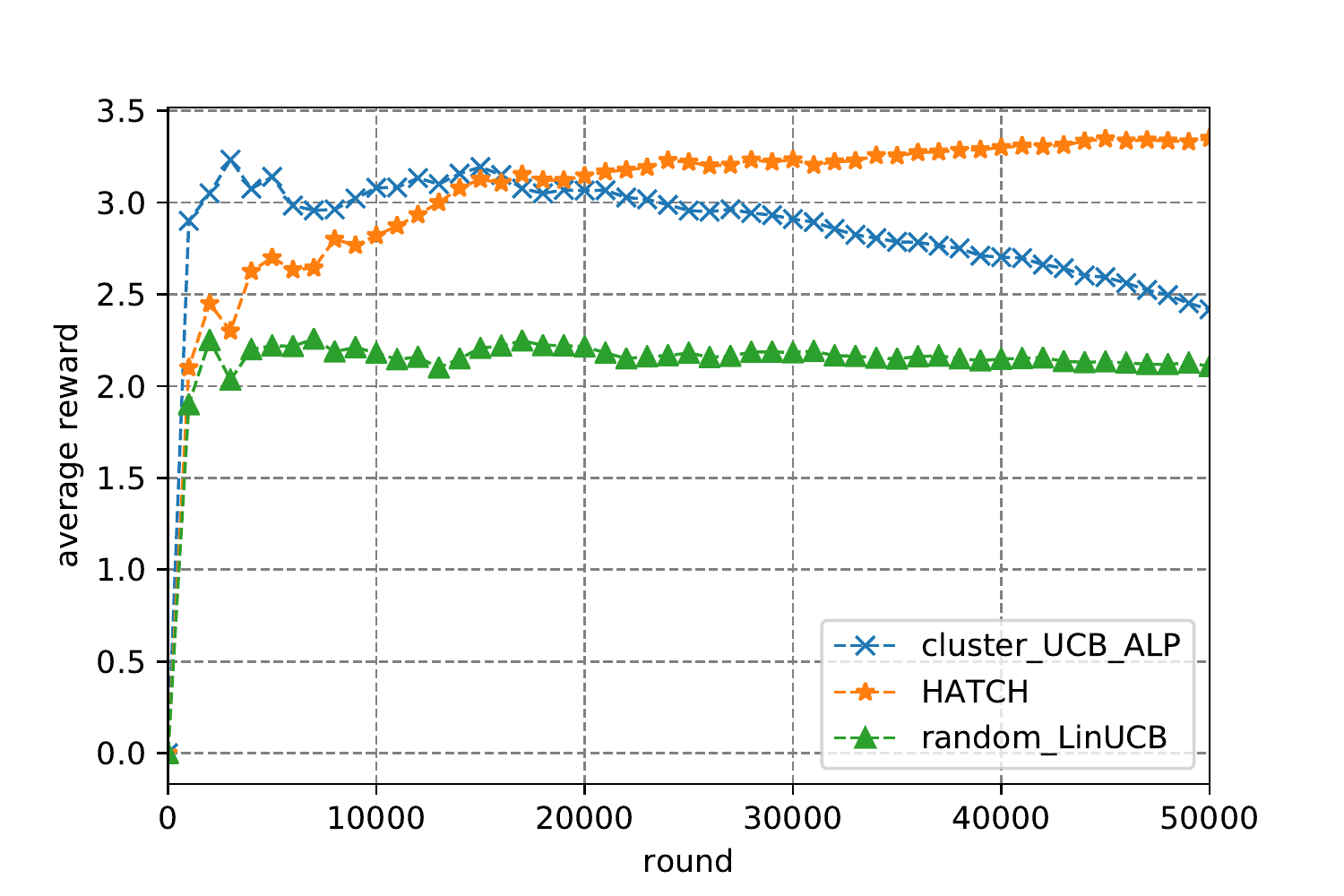}
% \caption{$\rho = 0.375$}
\end{minipage}%
}
\subfigure[$\rho = 0.5$]{
\begin{minipage}[t]{0.225\linewidth}
\centering
\includegraphics[width=1\textwidth]{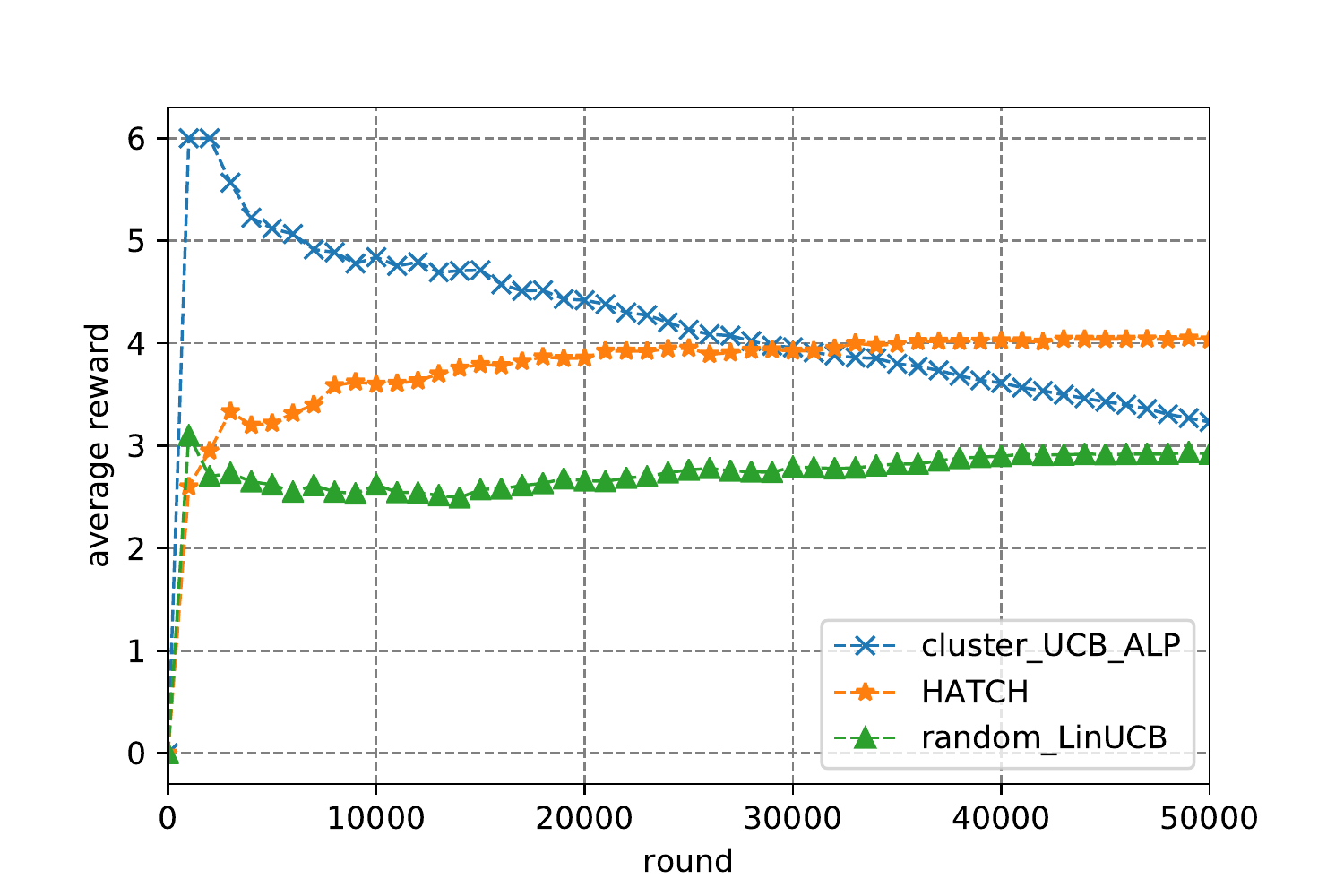}
% \caption{$\rho = 0.5$}
\end{minipage}
}
\caption{ Performance comparisons of averaged reward (CTR) among random LinUCB, cluster UCB-ALP, and proposed HATCH on Yahoo! Today module. The figure shows the trend of CTR for the different execution rounds. (a), (b), (c) and (d) illustrate different settings of resource ratio $\rho = \{0.125, 0.25, 0.375, 0.5\}$, respectively.}
\label{fig5}
\end{figure*}

\begin{figure*}
\centering
\subfigure[$\rho = 0.125$]{
\begin{minipage}[t]{0.225\linewidth}
\centering
\includegraphics[width=1\textwidth]{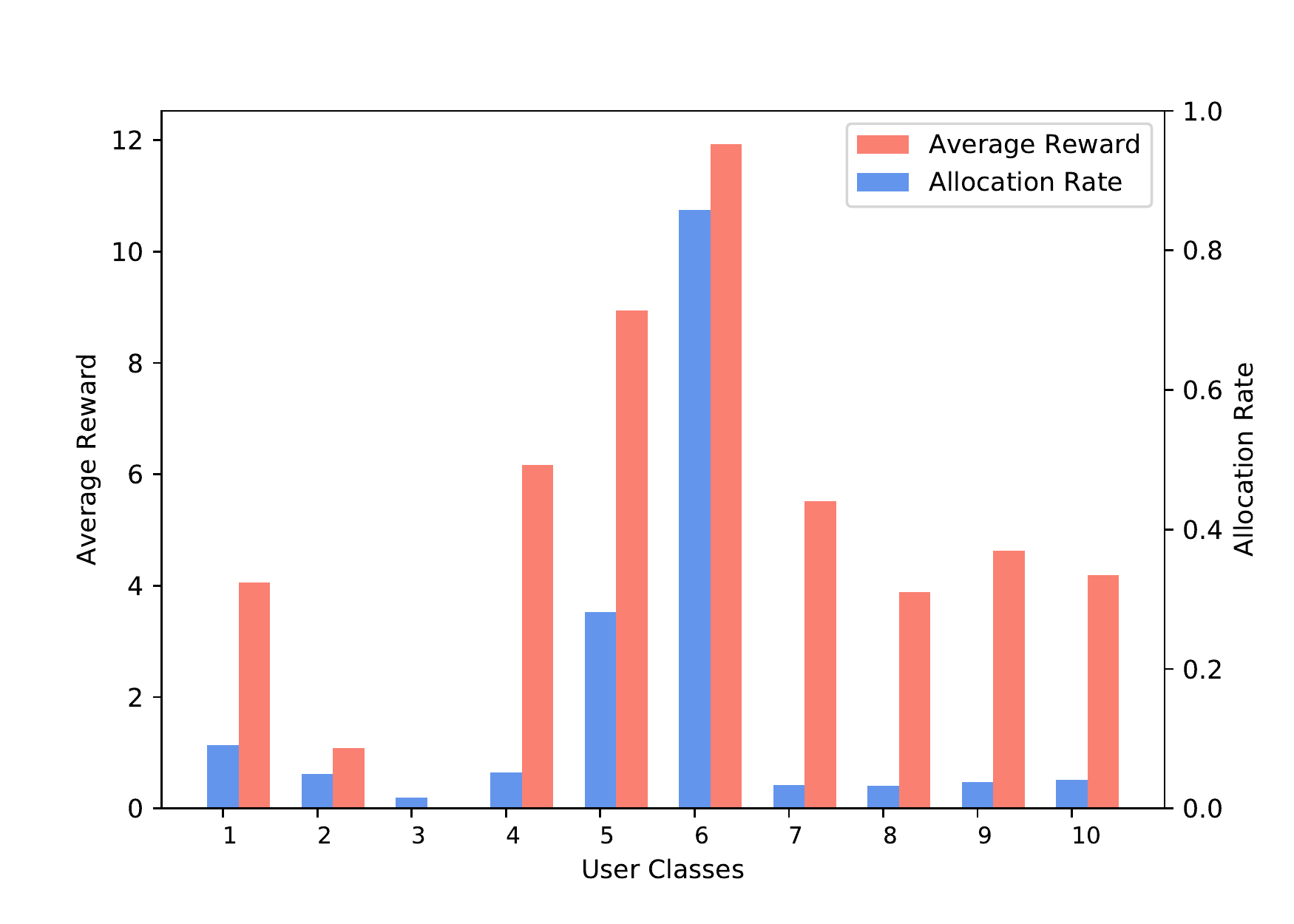}
\end{minipage}%
}
\subfigure[$\rho = 0.25$]{
\begin{minipage}[b]{0.225\linewidth}
\centering
\includegraphics[width=1\textwidth]{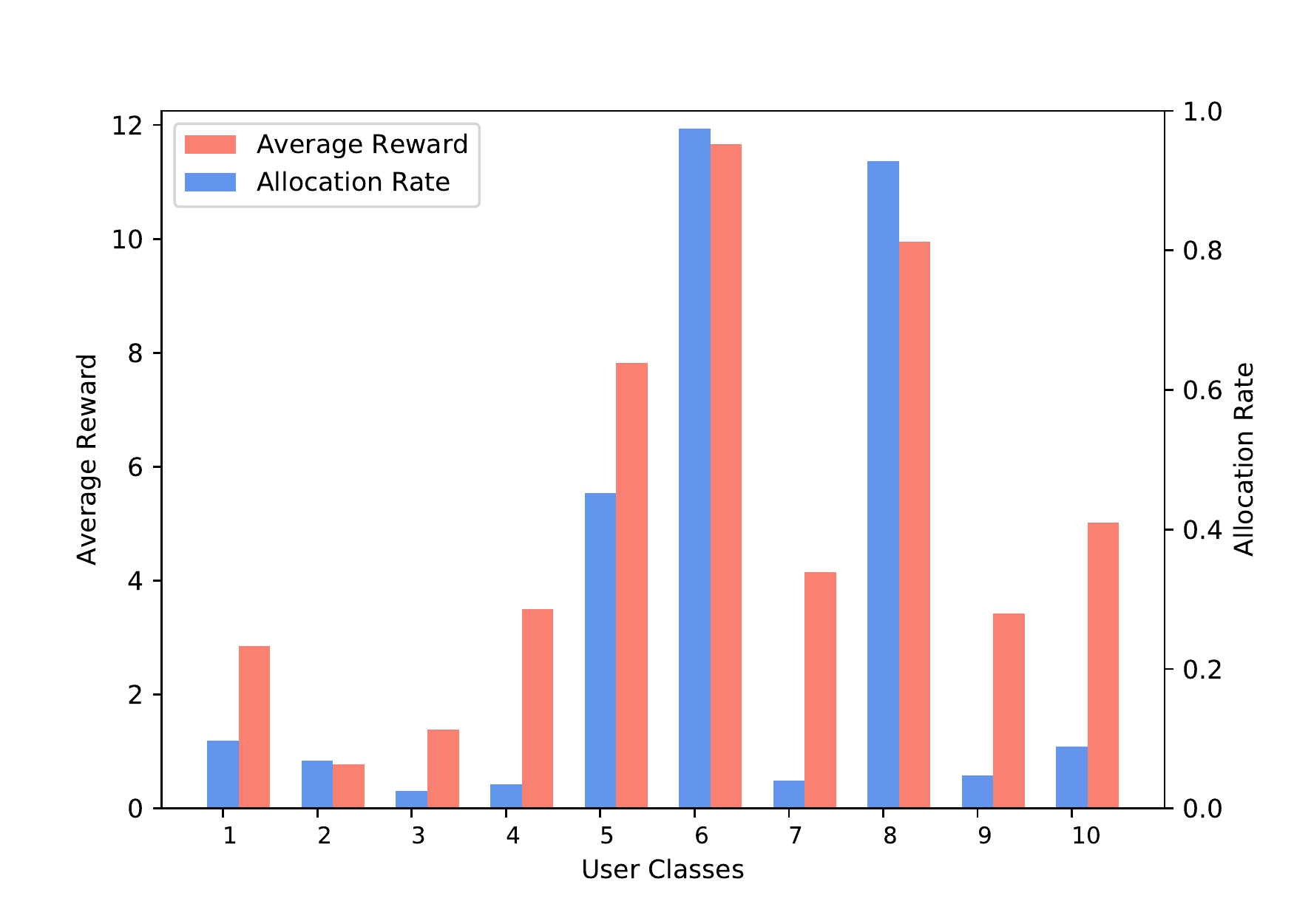}
\end{minipage}%
}
\subfigure[$\rho = 0.375$]{
\begin{minipage}[b]{0.225\linewidth}
\centering
\includegraphics[width=1\textwidth]{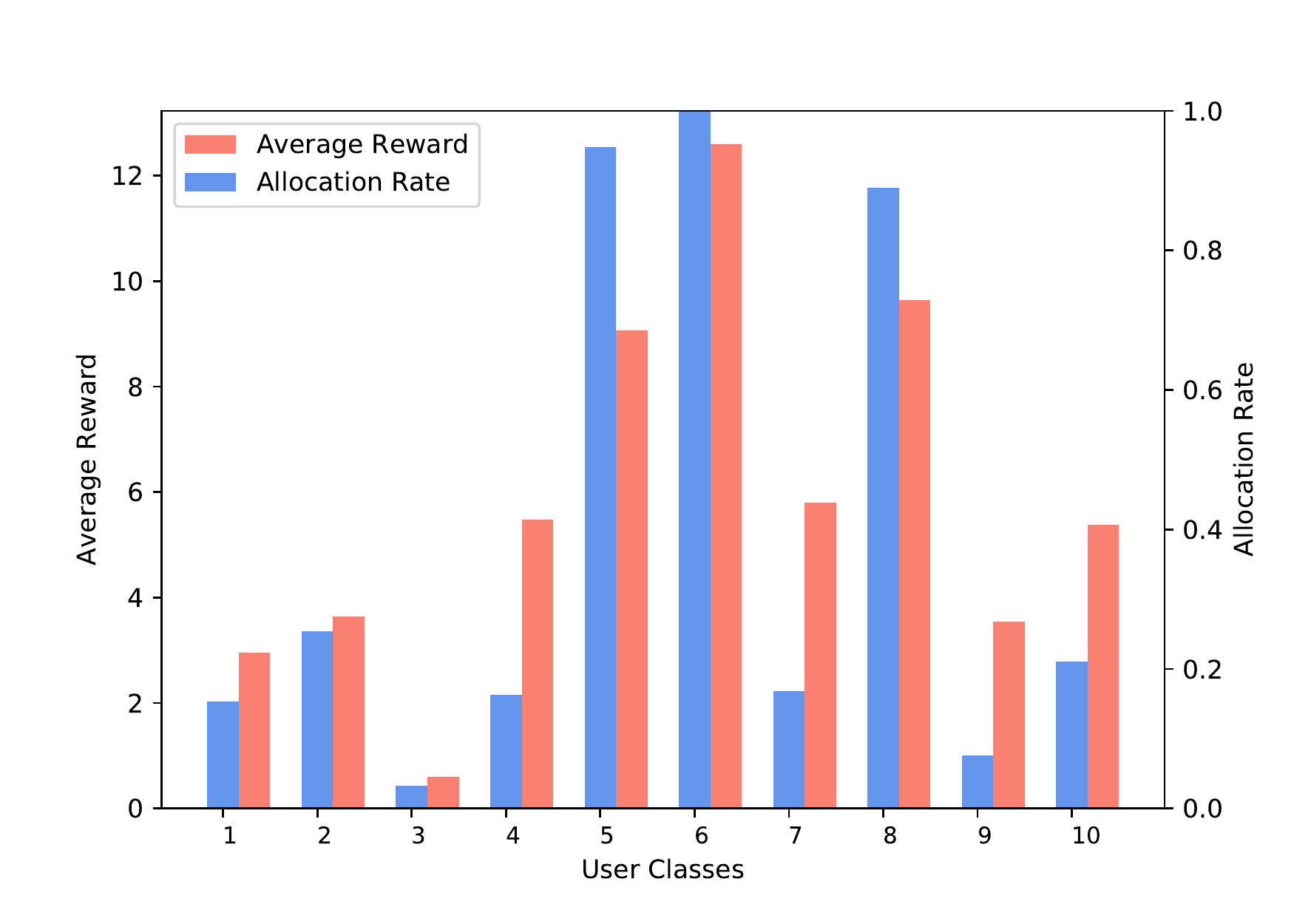}
\end{minipage}%
}
\subfigure[$\rho = 0.5$]{
\begin{minipage}[b]{0.225\linewidth}
\centering
\includegraphics[width=1\textwidth]{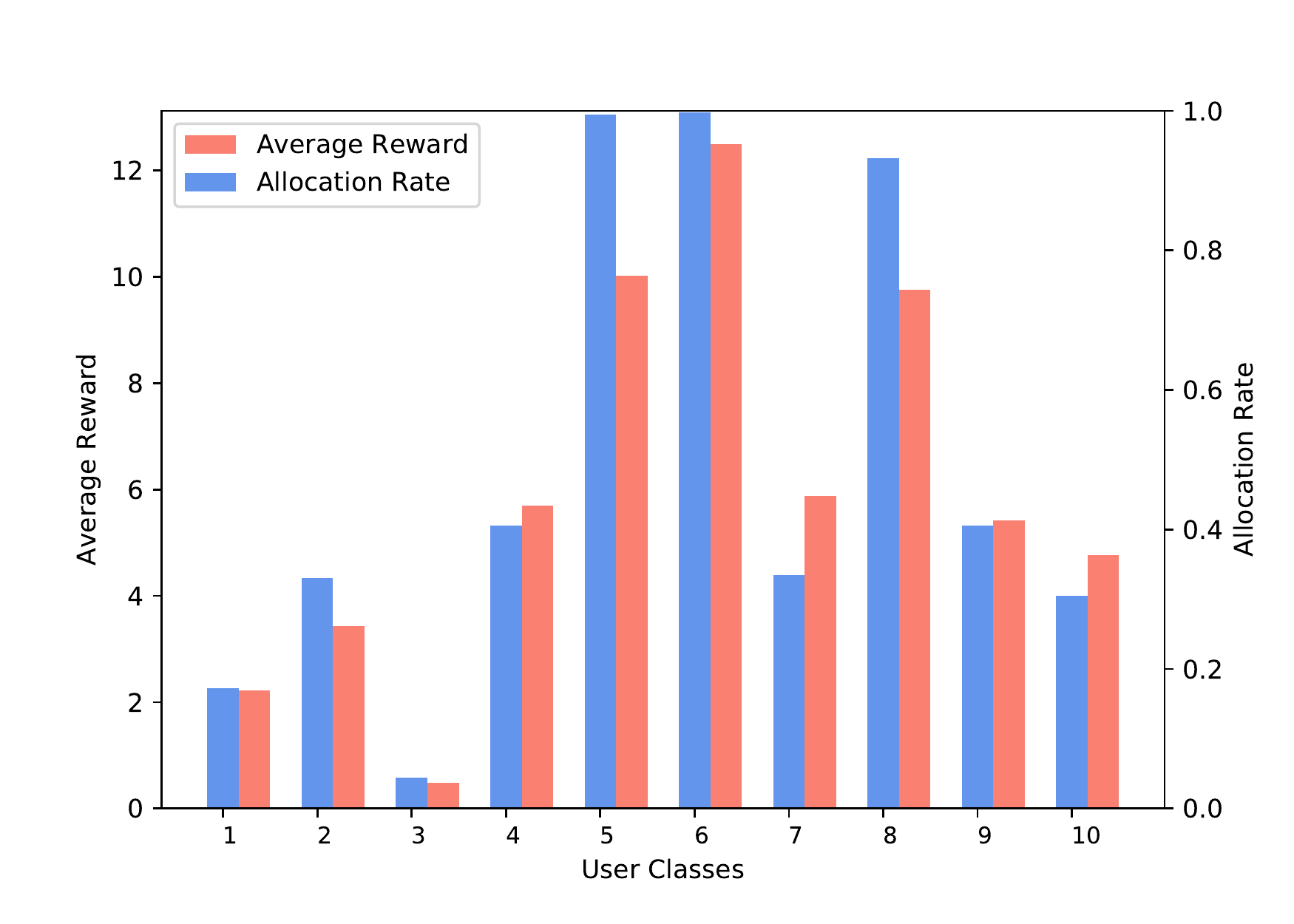}
\end{minipage}
}
\caption{ HATCH's statistic results of averaged reward and resource allocation rates for ten different user classes after executing 50000 rounds on Yahoo! Today module. (a), (b), (c) and (d) illustrate different settings of budget-time horizon ratio $\rho = \{0.125, 0.25, 0.375, 0.5\}$.}
\label{fig6}
\end{figure*}

\begin{table}
  \caption{The averaged rewards result (CTR) on Yahoo! Today Module after executing 50000 rounds.}
  \label{tab:freq}
  \begin{tabular}{ccccc}
    \toprule
    $\rho$&0.125&0.25&0.375&0.5\\
    \midrule
    greedy-LinUCB&0.83&1.69 & 2.49 & 3.29\\
    random-LinUCB&0.72& 1.54& 2.11&2.92\\
    cluster-UCB-ALP&0.82&  1.52 & 2.41 & 3.23 \\
    \textbf{HATCH}&\textbf{1.12}& \textbf{2.36}& \textbf{3.35}& \textbf{4.04}\\
  \bottomrule
\end{tabular}
\end{table}

The evaluation process is summarized as follows:
\begin{itemize}
    \item We map all the historical contextual user classes into $J$ buckets and initialize them as empty at the beginning.
    \item Put the context of user class $j$ into data bucket $bucket_j$. In each round, we sample a user class $j$ by the distribution $\phi$. 
    \item Then we sample data randomly from the $bucket_j$ and choose the arm recommended by the current bandit algorithm. If the selected arm is not the same as the arm in the sampled event, the evaluator will re-sample another event from $bucket_j$ until the selected arm is the same as the sampled arm. 
    \item Put the current event $(x_t, a_t, r_t)$ into historical data set $h_t$.
    \item Conduct the policy learning by using the data set.
\end{itemize}

\begin{algorithm}[t]
\caption{Evaluation from a static distribution}
\begin{algorithmic}[1]
\REQUIRE $\phi$, $\mathcal{G}, X, T>0, policy:p$
\STATE $J = \mathcal{G}(X)$
\STATE $h_0 = \emptyset$ \{An initially empty history\}
\STATE $R_0 = 0$ \{An initially zero total payoff\}
\STATE $Bucket = \{bucket_1, bucket_2,... bucket_J\}$

\FOR{$j = 1, 2,.. J$}
    \STATE {Put the $x$ whose class is $j$ into $bucket_j$}
\ENDFOR

\FOR{t = 1,2 ...T}
    \STATE sample a user class j via distribution $\phi$
    \REPEAT
        \STATE sample event $(x_t, a_t, r_t)$ from $bucket_j$
    \UNTIL{$p(h_{t-1},x)$ equals to $a_t$}
    \STATE $h_t  \leftarrow [h_{t-1}:(x_t,a_t,r_t)]$
    \STATE $R_t \leftarrow R_{t-1} + r_a$
    \STATE delete $(x_t, a_t, r_t)$ from $bucket_j$
\ENDFOR
\STATE Output: average reward $ = R_t/T$
\end{algorithmic}
\end{algorithm}

\textbf{Experiments on logging data set}
: We set a group of experiments on historical logging data. We compare our method (HATCH) with three baselines mentioned in synthetic experiments. Baselines are random-LinUCB, gready-LinUCB and cluster-UCB-ALP algorithm by using the evaluation strategy in Algorithm 2. The evaluate metric is to compare averaged reward (CTR) in this experiment.

\begin{table*}
  \caption{The occupancy rate of the user contexts among ten class after 50000 execution rounds.}
  \vspace{-3mm}
  \label{tab:freq1}
  \begin{tabular}{ccccccccccc}
    \toprule
    user class&class1&class2&class3&class4&class5&class6&class7&class8&class9&class10\\
    \midrule
    0.125&0.031&0.014 & 0.13 & 0.063&0.254&0.483 & 0.0464 & 0.288&0.0346 & 0.0306 \\
    
    0.25&0.017&0.010 & 0.12 & 0.021&0.207&0.262 & 0.027 & 0.391&0.021 & 0.032 \\

    0.375&0.018&0.023 & 0.009 & 0.063&0.292&0.184 & 0.080 & 0.255&0.022 & 0.052\\
    
    0.5&0.014&0.024 & 0.008 & 0.128&0.223&0.137 & 0.116 & 0.195&0.095 & 0.055 \\
  \bottomrule
\end{tabular}
\vspace{-2mm}
\end{table*}

%Since we do not know the user class distribution at first, some setting is not identical to synthetic experiments. We use GMM to evaluate distribution and here we set the number of clusters is 10. And we use the GMM model to predict the user cluster of every $x$ in round t. The baselines of UCB-ALP and HATCH have some differences with synthetic experiments as we illustrate following.

%\textbf{cluster-UCB-ALP:}  We cluster dataset into $J$ class corresponding to J user class. When agent observes a context, details of context will be ignored and the agent just records $j$-th user class's appearance. This setting can be used in UCB-ALP and cluster UCB-ALP.

%\textbf{HATCH:} We evaluate the HATCH by using the same cluster model as cluster-UCB-ALP.

For each experiment of the four methods, we set the round number as 50000 since the number of valid sample events (not dropped by the evaluation algorithm) of four methods are all around 70000 events. It is reasonable that we are able to evaluate all the comparison algorithms with 50000 rounds and all of the algorithms will have stable performance. 

In addition, we set the parameter $\alpha=1$ for random-LinUCB,  greedy-LinUCB as well as HATCH algorithm. For HATCH, we keep $\alpha$ as the same in both of resource allocation and personal recommendation level.

We show the trends of averaged reward (CTR) with different round number in Figure 3 and the result of all four methods after executing 50000 rounds in Table 1. 

From Figure 3 and the Table 1, we can observe that random LinUCB performs the worst with all different conditions of $\rho$. It is not efficient for this method to conduct the exploration in this setting since sometimes it might lose high value contexts which may benefit to the update of parameters. The performance of greedy-LinUCB can be seen in Table 1, it is better than the performance of random-LinUCB since it has fully explored the environment at the beginning. However, the averaged reward sharply declined when the resource is exhausted. Many of the previous works which adopted this kind of strategy will be not conducive to long-term reward with the constrained exploration resource even if the performance is a little bit better than random-LinUCB. When the environment changes, it might not be flexible to meet the change of reward function for the environment. The proposed HATCH outperforms the other methods in this experiment. The result shows that our algorithm tends to choose the higher valuable contexts to allocate the resource during the learning process, which means it is more reasonable for the exploration and exploitation trade off and it benefits for a longtime exploration in the dynamic algorithm. Cluster-UCB-ALP performs not worst in this setting, which illustrates that allocation strategy via linear programming is reasonable. However, it has the same issues as the experiments on synthetic data, without considering user contextual features, it could not utilize the preference of users for personalized recommendations.

To compare different kinds of resource setting, we set the resource constraint ratio $\rho = \{0.125, 0.25, 0.375, 0.5\}$, respectively. From Figure 2 and Table 1, the averaged reward of our algorithm performs better than other methods in different ratio settings. When $\rho$ is getting larger, the CTR of HATCH becomes higher. 

In Figure 4, we compare the resource allocation rates and the averaged reward distributions among different user classes to analyse the performance of resource allocation strategy and the policy exploration of HATCH. We also illustrate the occupancy rates of different user classes in Table 2, which are the normalized rates of the number of events that are allocated the resources. In Table 2, the occupancy rates are decided by the allocation rate and total number of event in each class. Figure 4 illustrates that the allocation rates have positive correlation with averaged rewards. When $\rho$ is smaller (such as $\rho = 0.125$), which means the allocation level has a larger ability of exploration, it has a higher probability to retain some contexts which might have lower expected value. In Figure 4(a), the averaged rewards of class 1 and class 8 are almost same, however, the allocation rate of class 1 is larger than that of class 8. It is because that the class 1 has less appearance than class 8, where we can observe that the occupancy rate (Table 2) in class 1 is less than it in class 8 when $\rho = 0.125$. It means there are less events which are used to update the parameters in class 1 than class 8 so that class 1 has more probability for exploration. 

Meanwhile, with the increasing of $\rho$, the resource allocation rate has increased and appears to be stable because HATCH has learned a more precise expected values for each user class in the allocation level, which means uncertainty is lower than that in smaller $\rho$. From Table 2 we can find that the occupancy rates of different user class become closer with $\rho$ increasing. The reason is that if we have more data to train a policy, the exploration of the allocation level will be decreased in the end while HATCH shows more tends of exploitation.
\section{Appendix}
\subsection{Proof of Lemma 1}

\textbf{Proof:}

Let $\sigma_{j,t} = (1+\widetilde{\alpha}) \widetilde{s}_{j,t}$,

\begin{equation}\label{eq7}
\hat{u}_{j,t}+\sigma_{j,t} \ge \hat{u}_{j',t}+\sigma_{j',t}\nonumber
\end{equation}
\begin{equation}
\begin{split}
[\hat{u}_{j,t} - (u_j + \sigma_{j,t})]+&[u_j-u_{j'}+ 2\sigma_{j,t}] \ge [\hat{u}_{j',t}-(u_{j'}-\sigma_{j',t})]\nonumber
\end{split}
\end{equation}
\begin{align}
    \mathbb{P}( \hat{u}_{j,t} \ge \hat{u}_{j',t}) &=  \mathbb{P}(\hat{u}_{j,t} - u_j \ge \sigma_{j,t}) \nonumber\\
    &+ \mathbb{P}(\hat{u}_{j',t}-u_{j'}\le -\sigma_{j',t})\nonumber \\ 
    &+ u_j+ 2\sigma_{j,t} \ge u_{j'}
\end{align}
In our setting, for each user class $j$, the text $\widetilde{s}_{j,t}$ is identical. Thus, $A_{j,t}$ is a symmetric matrix, and $det(A_{j,t}) = 1 + N_j(t-1)\|\widetilde{x}_j\|^2 \ge 1$, where $N_j(t-1)$ denotes the number of times that is allocated the resource until round $T$.

$A_{j,t}^{-1}$  can be decomposed as $A_{j,t}^{-1} = {A'}_{j,t}^\top{A'}_{j,t}$

According to the Cauchy-Schwarz:
\begin{align}
\widetilde{s}_{j,t} & = \sqrt{\|\widetilde{x}_{j}^\top {A}_{j,t}^{-1} \widetilde{x}_{j}\|}\nonumber\\
& = \sqrt{\|\widetilde{x}_{j}^\top{A'}_{j,t}^\top\|\|{A'}_{j,t}\widetilde{x}_{j}\|}\nonumber\\
& = \|\widetilde{x}_{j}\| /\sqrt{ (1+N_j(t-1)\|\widetilde{x}_{j}\|^2)}\nonumber\\
& < 1/\sqrt{N_j(t-1)}\label{eq2}
\end{align}
and according to the Hoeffding's inequality:
\begin{align}
\mathbb{P}(\hat{u}_{j,t} - u_j \ge \sigma_{j,t}) &\le exp(-2N_j(t-1)\sigma_{j,t}^2)=t^{-1}
\label{lemma1}
\end{align}
where $1+\widetilde{\alpha} = \sqrt{\frac{logt}{2}}$, then $\mathbb{P}(\hat{u}_{j',t}-u_{j'}\le -\sigma_{j',t})$ is the same as eq.\ref{lemma1}.

The remaining of the proof refers to the proof of Lemma 1 in \cite{jiang2013bandits}. When $N_j(t-1)\ge \lceil\frac{2logT}{(u_j-u_{j'})^2}\rceil$, we can get the following:  
\begin{align}
u_j+ 2\sigma_{j,t} \le u_j+2(u_{j'}-u_j)\sqrt{\frac{logt}{2logT}} \le u_{j'}\nonumber
\end{align}
means that 
\begin{align}
\mathbb{P}(u_j+ 2\sigma_{j,t} \ge u_{j'}) =0\nonumber
\end{align}
Then
\begin{equation}\label{eq6}
\mathbb{P}( \hat{u}_{j,t} \ge \hat{u}_{j',t}| N_j(t-1)\ge l_j) \le 2t^{-1}
\end{equation}

\subsection{Proof of Theorem 1}

\textbf{Proof of step 2:}

For $j \not= j'$, let $l_{j} = \frac{2logT}{(\Delta_j)^2}$ where $\Delta_j = inf\{|{u}_{j'}-{u}_j|\}$, where $j' \in J$.

\begin{align}
    &\mathbb{E}[N_j(T)] \le
    l_{j}+\sum\mathbb{P}\{\widetilde{x}_{t},N_j(T)\ge l_{j}\} \le
    l_{j}+\sum_{t = 1}^{T}2t^{-1}
\end{align}
Follow the facts that $\sum_{t = 1}^{T}t^{-1}\le1+\log T$. 

\textbf{Lemma 2 (referred from Theorem 3 in \cite{abbasi2011improved})} For $\forall t\ge 0$ and $ x_{t} \in \mathbb{R}$, $	\left \langle x_{t}^\top\theta_{t}\right \rangle\in[-1,1]$. Then, with probability at least $1-\delta$, the regret $R$ have the upper bound of 
\begin{equation}
    R_n\le4\sqrt{tlog(\lambda+t)}(\sqrt{\lambda} + \sqrt{2log(1/\delta)+log(1+t/\lambda)})
\end{equation}

Then according to Lemma $2$, $R^{(1)}(T,B)$ can be bounded as 
\begin{align}
    &R^{(1)}(T,B)\nonumber = [\sum_{j = 1}^{J}\sum_{t = 1}^{N_j}\sum_{a = 1}^{\mathcal{A}}x_{t}({\theta}_{t,j,a}^*-{\theta}_{j,a})]\nonumber\\
    &\le\sum_{j = 1}^{J}\sum_{t = 1}^{N_j}\sum_{a=1}^{\mathcal{A}}|x_t({\theta}_{t,j,a}-\theta_{j,a}^*)|\nonumber\\
    % &\le \sum_{j = 1}^{J}\mathbb{E}[N_j(T)]\sum_{a = 1}^{\mathcal{A}}\|x_t\|_{A^{-1}}(\sqrt{2log(\frac{det(A)^{\frac{1}{2}}}{\delta})}+\|\theta^*\|)\nonumber\\
    % & \le \sum_{j = 1}^{J}\mathbb{E}[N_j(T)]\sum_{a = 1}^{\mathcal{A}}\|x_t\|_{A^{-1}}(\alpha + 1)\nonumber\\
    & \le \sum_{j = 1}^{J}\beta_j(\sqrt{\mathbb{E}[N_j(T)]log(\lambda+\mathbb{E}[N_j(T)])})
    +\beta\sqrt{{B}log(\lambda+B)}\label{equ:2}
\end{align}
where 
$$\beta_j = \sqrt{\lambda}+\sqrt{2log(1/\delta) + log(1+2(logT/\Delta_j^2+logT+1)/\lambda))}$$
$$\beta = \sqrt{\lambda}+\sqrt{2log(1/\delta) + log(1+B/\lambda))}$$
\textbf{Proof of step 3:}
The process of the proof follows the UCB-ALP\cite{wu2015algorithms}. Different with UCB-ALP, the retain probability ($\hat{p}$) are calculate by global resource allocation level using user class context. Then we replaced $u_j$ in $v(\rho)$ by the best condition $u_j^*$.

\subsubsection{Non-boundary cases:}
We set $\varepsilon_0(t)$ to be the nearly correct order of user class. Then $v^*(\tau,b_\tau) = v(b_{\tau}/\tau)$. Let $\zeta = \frac{1}{2}min\{\rho-q_{\widetilde{j}(\rho)}, q_{\widetilde{j}(\rho)+1}-\rho\}$, if $b\in[\rho-\zeta, \rho + \zeta]$, then

\begin{align}
&\mathbb{E}[\Delta v_{\tau}, \varepsilon_0(T-\tau+1)] =\nonumber \\ &\sum_{b = 0}^{B} \mathbb{E}[\Delta v_{\tau}| \varepsilon_0(T-\tau+1)] \mathbb{P}\{b_\tau = b, \varepsilon_{0}(T-\tau+1)\}
\end{align}
where
\begin{align}
    \mathbb{E}[\Delta v_{\tau}, \varepsilon_0(T-\tau+1)] = v(\rho)-v(b/\tau)
\end{align}
\begin{align}
&\mathbb{P}\{b_\tau = b, \varepsilon_{0}(T-\tau+1)\} = \mathbb{P}\{b_\tau = b\} - \sum_{s = 1}^{2}\mathbb{P}\{b_\tau = b,\varepsilon_s(T-\tau+1)\}
\end{align}
\begin{align}
\text{Then \quad}
    &\mathbb{E}[\Delta v_{\tau}, \varepsilon_0(T-\tau+1)]  \nonumber\\
    &\le[u_1^*-u_J^*+2\bar{u}^*]e^{-2\zeta^2\tau}+[\bar{u}^*\sum_{s = 1}^{2}\mathbb{P}\{\varepsilon_s(T-\tau+1)\}]\nonumber\\
    &\le[u_j-u_1+2C+2\bar{u}]e^{-2\zeta^2\tau}+[\bar{u}+C]\sum_{s = 1}^{2}\mathbb{P}\{\varepsilon_s(T-\tau+1)\}]
\end{align}
When it happened the wrong ranking case, for $1\le s\le2$, there existed $\Delta_\tau\le v(\rho)$ in any possible ranking results. Then it can obtain the following equation:
\begin{align}
    \mathbb{E}[\Delta v_\tau,\varepsilon_s(T-\tau +1)] \le \nonumber
    v(\rho)\mathbb{P}[\varepsilon_s(T-\tau+1)]
\end{align}
Since $R^{(2)}(T,B) = \sum_{\tau = T}^1\mathbb{E}[\Delta v_{\tau}]$, then we have:
\begin{align}
&R^{(2)}(T,B)\nonumber\\ &\le\frac{[u_j-u_1+2C+2\bar{u}]e^{-2\zeta^2\tau}}{1-e^{-2\zeta^2\tau}}\nonumber\\
&+[\bar{u}+v(\rho)+C]\sum_{s = 1}^{2}\mathbb{E}[T^{(s)}]
\end{align}
where $T^{(s)} = \sum_{t = 1}^T\mathbb{I}(\varepsilon_s(t))(s = 1,2)$ is the type-$s$ ranking error.

From \cite{wu2015algorithms}, we have the following equation in our setting:
\begin{align}
    \mathbb{E}[T^{(1)}] \le \frac{2e^{-2\zeta^2}}{1-2e^{-2\zeta^2}} +\sum_{j = 1}^{\widetilde{j}(\rho)}\mathbb{E}[C_j^{(1)}(T)]
\end{align}

Let $\hat{l}_j = \frac{2logT}{g_j(1-\xi)\xi^2(\Delta_j)^2}$ where $g_j = min\{\phi_j,\zeta\}$ and $\xi\in (0,1)$, according to Lemma 5 in \cite{wu2015algorithms} and our Lemma 1 we have: 
\begin{align}
    \mathbb{E}[C_j^{(1)}(T)]\le \hat{l}_{\widetilde{j}(\rho)+1} + \sum_{t = 1}^{T}(2t^{-1}+T^{-4})
\end{align}
Let $\xi = \frac{2}{3}$, then
\begin{align}
    \mathbb{E}[T^{(1)}] \le \sum_{j = 1}^{\widetilde{j}(\rho)}\frac{27logT}{2g_{\widetilde{j}(\rho)+1}(\Delta_{{\widetilde{j}(\rho)+1}})^2}+2\widetilde{j}(\rho)logT
\end{align}
The proof of $\mathbb{E}[T^{(2)}]$ is same as the $\mathbb{E}[T^{(1)}]$, we have the following:
\begin{align}
    \mathbb{E}[T^{(2)}] \le \sum_{j = \widetilde{j}(\rho)+2}^{J}\frac{27logT}{2g_j(\Delta_j)^2}+2\widetilde{j}(\rho)logT
\end{align}
Thus, we could obtain the final proof result for non-boundary cases:
\begin{align}
    &\limsup_{t\rightarrow\infty}\frac{R^{(2)}(T,B)}{logT}\nonumber\\
\le   &[\bar{u} + C + v(\rho)]
[\sum_{j = 1}^{\widetilde{j}(\rho)}\frac{27}{2g_{\widetilde{j}(\rho)+1}[\Delta_{\widetilde{j}(\rho)+1}]^2}\nonumber \\
+ &\sum_{j = \widetilde{j}(\rho)+2}^{J}\frac{27}{2g_j[\Delta_{j}]^2}+ 2J]\label{eq4}
\end{align}

\subsubsection{Boundary cases}
Let $\zeta = \frac{1}{2}min\{\rho-q_{\widetilde{j}(\rho)-1}, q_{\widetilde{j}(\rho)+1}-\rho\}$. When $\varepsilon_0$ occurs and $b/\tau\in [\rho-\zeta,\rho+\zeta]$, there existed $\Delta v_\tau\le(u_1-u_J)|\rho-b/t|$, then we can have
\begin{align}
    &\mathbb{E}[\Delta v_{\tau}, \varepsilon_0(T-\tau+1)] \nonumber\\
    &\le (u_1+C)\mathbb{E}[|b/\tau-\rho|]+v(\rho)\sum_{b\notin[\rho-\zeta,\rho+\zeta]}\mathbb{P}\{b_\tau = b\}\nonumber\\
    &\le (u_1+C)\sqrt{\frac{Var(b_\tau)}{\tau^2}} + 2v(\rho)e^{-2\xi^2\tau s}
\end{align}
where 
\begin{align}
    \sum_{\tau =  1}^{T}\sqrt{\frac{Var(b_\tau)}{\tau^2}} 
    &=\sum_{\tau =  1}^{T}\sqrt{\frac{(T-\tau)\rho(1-\rho)}{(T-1)\tau}}\nonumber\\
    & \le \sqrt{\rho(1-\rho)}\sum_{\tau =  1}^{T}\sqrt{\frac{1}{\tau}} \nonumber \\ 
    &\le 2\sqrt{\rho(1-\rho)}\sqrt{T}
\end{align}

If it occurred the ranking error situations for $s = 1,2$, we have the following equation:

\begin{align}
    &\mathbb{E}[\Delta v_{\tau}, \varepsilon_s(T-\tau+1)]\le v(\rho)\mathbb{P}\{\varepsilon_s(T-\tau+1)\}
\end{align}

By extending the lemma 6 in \cite{wu2015algorithms}, $R^{(2)}(T,B)$ can be divided into two parts, $R^{(2)}(T,B) = R^{(2)}_1(T,B)+R^{(2)}_2(T,B)$. Then we have the following proof in the boundary cases:
\begin{align}
    &R^{(2)}_1(T,B)\le  (u_1+C)\sqrt{\frac{Var(b_\tau)}{\tau^2}}\ + 2v(\rho)e^{-2\zeta^2\tau}\label{eq5}
\end{align}

\begin{align}
    &\limsup_{t\rightarrow\infty}\frac{R^{(2)}_2(T,B)}{logT}\nonumber\\
\le   &[\bar{u} + C + v(\rho)]
[\sum_{j = 1}^{\widetilde{j}(\rho)}\frac{27}{2g_{\widetilde{j}(\rho)+1}[\Delta_{\widetilde{j}(\rho)+1}]^2}\nonumber \\
+& \sum_{j = \widetilde{j}(\rho)+1}^{J}\frac{27}{2g_j[\Delta_{j}]^2}+ 2J]\label{eq4}
\end{align}

\section{Conclusions}
In this paper, we proposed HATCH, a linear contextual multi-armed bandits with dynamic resource allocation of recommendation, to solve resources constraint problem in recommendation system. Experimental results have shown that our algorithm achieves (i) higher performance than aforementioned methods, (ii) better adaptation and more reasonable resources allocation. We present and give a theoretical analysis to prove that HATCH achieves a regret bound of $O(\sqrt{T})$. For the future work, we notice that applying a linear contextual multi-armed bandits in real world setting is fully stop. There are several directions such as improving the robustness and adapting to environment changes or extend this method in to more general conditions.

\bibliographystyle{ACM-Reference-Format}
\bibliography{yang2020www}

\appendix

\end{document}